\documentclass[journal]{IEEEtran}

\usepackage{amsmath,amsfonts}
\usepackage{algorithmic}
\usepackage{algorithm}
\usepackage{array}
\usepackage[caption=false,font=footnotesize,labelfont=rm,textfont=rm]{subfig}
\usepackage{textcomp}
\usepackage{stfloats}
\usepackage{url}
\usepackage{verbatim}
\usepackage{graphicx}
\usepackage{cite}

\usepackage{multirow}
\usepackage{multicol}
\usepackage{booktabs}
\usepackage[section]{placeins}
\usepackage{color}
\usepackage{threeparttable}

\newcommand{\Rmnum}[1]{\expandafter\@slowromancap\romannumeral #1@}

\usepackage{makecell}

\usepackage{diagbox}
\usepackage[colorlinks=true, citecolor=blue, linkcolor=blue, urlcolor=black]{hyperref}

\usepackage{bm}
\usepackage{amssymb}
\usepackage{amsthm}
\usepackage{pifont}
\usepackage[table]{xcolor}
\usepackage{tikz}
\usepackage{orcidlink} %
\usepackage{arydshln}
\newcommand{\equref}[1]{Eq.~\textcolor{blue}{(\ref{#1})}}

\newtheorem{theorem}{Theorem}
\newtheorem{assumption}{Assumption}

\newcommand{\xmark}{\ding{55}} 

\definecolor{darkred}{rgb}{0.8, 0, 0}    
\definecolor{darkgreen}{rgb}{0, 0.5, 0}  
\definecolor{my_color}{HTML}{E8F3F1}

\newcommand{\colorcellwithFoneRange}[4]{%
  \pgfmathsetmacro{\precision}{#1/100}
  \pgfmathsetmacro{\recall}{#2/100}
  \pgfmathsetmacro{\denominator}{\precision + \recall}
  \pgfmathsetmacro{\fOne}{
    ifthenelse(\denominator==0, 0, 2 * \precision * \recall / \denominator)
  }
  \pgfmathsetmacro{\range}{#4 - #3}
  \pgfmathsetmacro{\normalizedFOne}{
    ifthenelse(\range==0, 0, (\fOne - #3)/\range)
  }
  \pgfmathsetmacro{\percent}{max(0, min(1, \normalizedFOne))}
  \pgfmathsetmacro{\red}{0.867 - 0.259*\percent}
  \pgfmathsetmacro{\green}{0.922 - 0.161*\percent}
  \pgfmathsetmacro{\blue}{0.969 - 0.067*\percent}
  \edef\cellcolorval{\noexpand\cellcolor[rgb]{\red,\green,\blue}}%
  \cellcolorval
}
\newcommand{\fOneCellRange}[4]{%
  \colorcellwithFoneRange{#1}{#2}{#3}{#4}%
  \pgfmathsetmacro{\precision}{#1/100}
  \pgfmathsetmacro{\recall}{#2/100}
  \pgfmathsetmacro{\denominator}{\precision + \recall}
  \pgfmathsetmacro{\fOneVal}{
    ifthenelse(\denominator==0, 0, 2 * \precision * \recall / \denominator)
  }
  \pgfmathsetmacro{\fOneRounded}{round(100*\fOneVal)/100}
  \pgfmathsetmacro{\maxRounded}{round(100*#4)/100}
  \ifdim \fOneRounded pt = \maxRounded pt
    \textbf{#1/\ #2}%
  \else
    #1/\ #2%
  \fi
}

\begin{document}

\title{ZeroPur: Succinct Training-Free Adversarial Purification}

\author{Erhu Liu$^{\orcidlink{0009-0003-7368-364X}}$, Zonglin Yang, Bo Liu$^{\orcidlink{0000-0002-3164-6299}}$,~\IEEEmembership{Member,~IEEE}, 
Xianjia Meng$^{\orcidlink{0000-0003-4485-457X}}$,
Xiuli Bi$^{\orcidlink{0000-0003-3134-217X}}$, Junwei Han$^{\orcidlink{0000-0001-5545-7217}}$,~\IEEEmembership{Fellow,~IEEE}, and Bin Xiao$^{\orcidlink{0000-0001-8469-5302}}$

\thanks{This work was supported in part by the National Science Foundation of China Joint Key Project under Grants U24B20173 and U24B20182, and by the National Natural Science Foundation of China under Grants 62576060 and 62376046. Additional support was provided by the Natural Science Foundation of Chongqing under Grant CSTB2023NSCQ-MSX0341 and the Science and Technology Research Program of the Chongqing Municipal Education Commission~(Grant No. KJQN202200635). 
\emph{(Corresponding author: Xianjia Meng; Xiuli Bi.)}}

\thanks{Erhu Liu, Bo Liu, Xiuli Bi and Bin Xiao are with the Department of Computer Science and Technology,  Chongqing University of Posts and Telecommunications, Chongqing 400065, China (e-mail: d240201026@stu.cqupt.edu.cn; boliu@cqupt.edu.cn; bixl@cqupt.edu.cn;
xiaobin@cqupt.edu.cn).}

\thanks{Zonglin Yang is with the Department of Biological Systems Engineering, University of Nebraska-Lincoln, NE 68583-0726, USA (e-mail: zyang41@unl.edu).}

\thanks{Xianjia Meng is with the School of Computer Science, Northwest University, Xi’an 710127, China (e-mail: xianjiam@nwu.edu.cn).}

\thanks{Junwei Han is with the School of Automation, Northwestern Polytechnical University, Xi’an 710072, China (e-mail: junweihan2010@gmail.com).}
 }

\markboth{Journal of \LaTeX\ Class Files,~Vol.~14, No.~8, August~2021}%
{Shell \MakeLowercase{\textit{et al.}}: A Sample Article Using IEEEtran.cls for IEEE Journals}

\maketitle

\begin{abstract}
 Adversarial purification is a kind of defense technique that can defend against various unseen adversarial attacks without modifying the victim classifier. Existing methods often depend on external generative models or cooperation between auxiliary functions and victim classifiers. However, retraining generative models, auxiliary functions, or victim classifiers relies on the domain of the fine-tuned dataset and is computation-consuming. In this work, we suppose that adversarial images are outliers of the natural image manifold, and the purification process can be considered as returning them to this manifold. Following this assumption, we present a simple adversarial purification method without further training to purify adversarial images, called ZeroPur. ZeroPur contains two steps: given an adversarial example, Guided Shift obtains the shifted embedding of the adversarial example by the guidance of its blurred counterparts; after that, Adaptive Projection constructs a directional vector by this shifted embedding to provide momentum, projecting adversarial images onto the manifold adaptively. ZeroPur is independent of external models and requires no retraining of victim classifiers or auxiliary functions, relying solely on victim classifiers themselves to achieve purification. Extensive experiments on three datasets (CIFAR-10, CIFAR-100, and ImageNet-1K) using various classifier architectures (ResNet, WideResNet) demonstrate that our method achieves state-of-the-art robust performance. The source code is publicly available at: \url{https://github.com/erhul/ZeroPur}.
\end{abstract}

 
\section{Introduction}
\IEEEPARstart{O}{ver} the past decade, a substantial body of work has confirmed that deep neural networks (DNNs) are highly vulnerable to imperceptible adversarial perturbations that cause wrong decisions~\cite{szegedy2013intriguing},~\cite{goodfellow2014explaining},~\cite{moosavi2016deepfool},~\cite{xie2019improving},~\cite{zhang2020principal},~\cite{croce2020reliable}. This potential vulnerability, despite their remarkable performance, poses a significant challenge for security-critical applications. Thus, designing effective and efficient adversarial defense techniques is essential to ensure the reliability and trustworthiness of DNNs in real-world scenarios.

One kind of adversarial defense technique is \textit{adversarial training}~\cite{madry2017towards},~\cite{zhang2019theoretically},~\cite{addepalli2022efficient},~\cite{jin2025s2o}, which involves adversarial examples in the model training, enabling the model to adapt adversarial perturbations empirically. However, these approaches typically require huge computational resources~\cite{shafahi2019adversarial},~\cite{wu2022towards} and suffer from performance degradation~\cite{laidlaw2020perceptual} in the presence of unseen attacks that are not involved in training. This limitation hinders the application of adversarial training in realistic scenarios.

Different from adversarial training, \textit{adversarial purification},~\cite{mao2021adversarial},~\cite{shi2021online},~\cite{zhang2021defense},~\cite{yoon2021adversarial},~\cite{nie2022diffusion},~\cite{hwang2023aid} aims to remove adversarial perturbations in adversarial examples. These methods do not require adversarial examples in the model training and effectively defend against unseen attacks, making them more applicable in real-world scenarios. However, the existing purification methods often depend on external generative models~\cite{yoon2021adversarial},~\cite{nie2022diffusion} or cooperation between external auxiliary functions and the victim classifier~\cite{mao2021adversarial},~\cite{shi2021online},~\cite{hwang2023aid}. Retraining generative models, parameterized auxiliary functions, or the classifier relying on the domain of the fine-tuned dataset is computationally demanding and restricts their flexibility.

\begin{figure*}
\centering\includegraphics[scale=0.75]{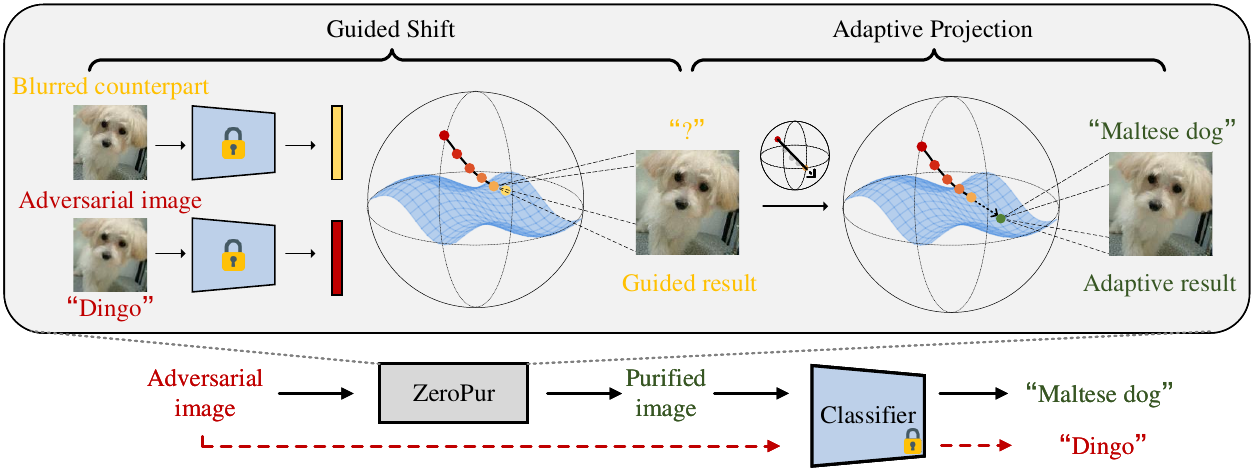}
\caption{An illustration of the proposed ZeroPur framework. ZeroPur consists of two stages: Guided Shift (GS) and Adaptive Projection (AP). In the GS stage, adversarial samples are iteratively shifted toward their blurred counterparts to obtain embeddings that are closer to the natural image manifold. In the AP stage, a directional vector is constructed based on the shifted embedding, and the adversarial sample is adaptively projected back onto the manifold, yielding the purified result. Despite its simple design, ZeroPur effectively purifies adversarial images, enabling the classifier to recover the correct prediction.}

 \label{workflow}
\end{figure*}

Inspired by the natural image manifold hypothesis, we suppose that adversarial images are outliers of the natural image manifold, and the purification process aims to return them to this manifold. We propose a simple adversarial purification method called ZeroPur, where ‘Zero' means our method does not need to train any external models, parameterized auxiliary functions, and victim classifiers. As illustrated in \figurename~\ref{workflow}, ZeroPur comprises two stages: \textbf{G}uided \textbf{S}hift (GS) and \textbf{A}daptive \textbf{P}rojection (AP) for purifying the adversarial image. Specifically, we observe that the blurring operator provides a reasonable direction towards natural image on the manifold for an adversarial sample. Therefore, GS iteratively guides adversarial sample towards its blurred counterparts, obtaining shifted embedding that approximate the manifold. Due to the inherent limitations of low-quality embeddings from blurred counterparts, adversarial sample may not precisely return to the target location, but the shifted embedding after multiple iterations provide a more reliable direction. Subsequently, we introduce AP to construct a directional vector based on this shifted embedding and project the adversarial image adaptively onto the manifold by maximizing the projection along this direction. Despite its embarrassingly simple design, ZeroPur consistently outperforms most state-of-the-art (SOTA) adversarial training and purification methods.

The main contributions of the current work are as follows:
\begin{itemize}
	\item We analyze the relationship between adversarial attack and adversarial purification based on the natural image manifold hypothesis, and show that a simple blurring operator can bring adversarial examples that are out of the natural image manifold closer to the manifold.
	\item We present a succinct adversarial purification approach named ZeroPur, including two stages: Guided Shift and Adaptive Projection, which requires no retraining of external models, parameterized functions, and the victim classifier.
	\item Extensive experiments demonstrate that the proposed approach outperforms most state-of-the-art auxiliary-based adversarial purification methods and achieves competitive performance compared to other external model-based purification methods.
\end{itemize}

 \section{Related Works}
 \label{sec:related_work}
In this section, we review representative defenses for adversarial robustness, focusing on adversarial training and adversarial purification. We further classify purification methods into external model-based and auxiliary-based paradigms. Their respective trade-offs in robustness, efficiency, and practicality motivate the design of our training-free purification framework.

\subsection{Adversarial training (AT)}
Since the discovery of adversarial examples, numerous defense strategies have been proposed to improve the robustness of DNNs. Among them, adversarial training (AT)~\cite{madry2017towards} has been widely recognized as one of the most effective empirical defenses~\cite{athalye2018obfuscated}. AT~\cite{madry2017towards},~\cite{zhang2019theoretically},~\cite{addepalli2022efficient},~\cite{jin2025s2o} improves robustness by integrating adversarial examples into training and reformulating the optimization objective.
Despite its effectiveness, AT suffers from several intrinsic limitations. It often degrades natural accuracy due to the accuracy--robustness trade-off~\cite{zhang2019theoretically} and may experience robust overfitting during later training stages~\cite{rice2020overfitting,wang2023balance}. Moreover, generating adversarial examples incurs substantial computational overhead and may introduce catastrophic overfitting~\cite{wong2020fast,jia2022boosting}. In addition, AT remains vulnerable to previously unseen attacks that were not included during training~\cite{laidlaw2020perceptual},~\cite{kaufmann2019testing},~\cite{dai2022formulating}.

In parallel with AT, early defenses also explored input preprocessing strategies~\cite{guo2017countering},~\cite{xie2017mitigating}, which attempt to suppress adversarial perturbations through fixed transformations in pixel space. However, Athalye~\cite{athalye2018obfuscated} showed that such methods rely on gradient obfuscation and fail to provide genuine robustness. Although ZeroPur also employs blur, it treats blur as a manifold-oriented guidance signal rather than the purification mechanism itself, fundamentally differing from conventional one-shot pixel-space preprocessing defenses.

\subsection{External model-based adversarial purification (EBP)}
Samangouei~\cite{samangouei2018defense} proposes defense-GAN, a generator that models the distribution of natural images, which enables the transformation from adversarial examples to natural images. Song~\cite{song2017pixeldefend} assumes that adversarial examples primarily reside in the low probability density region of the training distribution, and design PixelDefend to approximate this distribution using the PixelCNN~\cite{van2016pixel}. Zhou~\cite{zhou2021towards} introduces the Adversarial Noise Removal Network (ARN) framework, designed to learn attack-invariant features and recover natural samples from adversarial images. Later, utilizing score-based models~\cite{yoon2021adversarial} and diffusion models~\cite{nie2022diffusion},~\cite{song2024mimicdiffusion},~\cite{lee2023robust},~\cite{kang2023diffattack} as purification models is introduced and shown to achieve significantly improved robust performance. Recently, Lin~\cite{lin2024adversarial} proposes a framework called AToP that fine-tunes the purification model adversarially to integrate the benefits of both adversarial training and adversarial purification. However, these works rely on external generative models, substantial datasets (e.g., crafting adversarial examples), and computational resources.

\subsection{Auxiliary-based adversarial purification (ABP)}
In contrast to the above two approaches, ABP tends to introduce an auxiliary function to cooperate with the classifier to purify adversarial images. Shi~\cite{shi2021online} proposes a lightweight purification framework, SOAP, which employs self-supervised objectives to achieve online purification. Although SOAP removes the dependency on external generative models, it requires classifier retraining by incorporating auxiliary supervision. To further reduce purification overhead, recent studies introduce parameterized auxiliary functions and perform purification through dedicated lightweight modules. These auxiliary functions are designed to be computationally cheaper than classifiers while maintaining purification effectiveness. For example, Mao~\cite{mao2021adversarial} introduces a two-layer network to estimate contrastive features for adversarial purification, while Hwang~\cite{hwang2023aid} proposes AID-Purifier, an auxiliary discriminator based on information maximization principles to transform adversarial images into natural images. 
Although ZeroPur is categorized as an auxiliary-based purification method in our paper, it differs from conventional ABP approaches~\cite{shi2021online},~\cite{mao2021adversarial},~\cite{hwang2023aid} by requiring no auxiliary training. Instead, ZeroPur directly leverages fixed classifier representations for purification, making it a training-free instantiation of ABP.

\section{Methodology }
\label{sec:method}
Our contributions are twofold: (1) a training-free purification framework that projects adversarial examples toward the natural image manifold using only the victim classifier; (2) a two-stage purification strategy, Guided Shift and Adaptive Projection, for progressive adversarial restoration. The resulting method improves robustness against unseen attacks without retraining or auxiliary models.

\begin{figure*}[t]
\centering
    \begin{minipage}[c]{0.795\textwidth}
        \includegraphics[width=\textwidth]{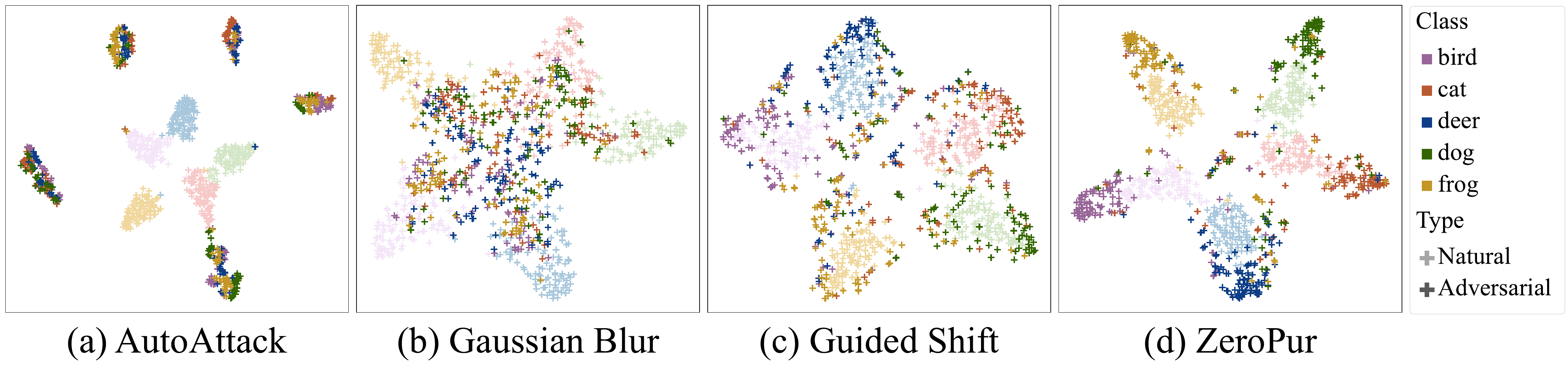}
    \end{minipage}
    \begin{minipage}[c]{0.002\textwidth}
        \centering
        \textcolor{gray}{
            \rotatebox{90}{
                \texttt{·············}
            }
        }
    \end{minipage}
    \hspace{0.002\textwidth}
    \begin{minipage}[c]{0.18\textwidth}
        \includegraphics[width=\textwidth]{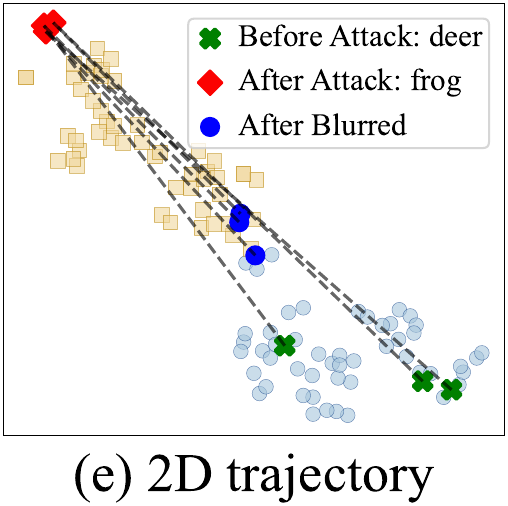}
    \end{minipage}
    
\caption{ The impact of adversarial samples and their improvement through different operations, visualized using t-SNE~\cite{maaten2008visualizing}. (a) AutoAttack~\cite{croce2020reliable} causes adversarial samples to deviate from the manifold; (b) Gaussian Blur, (c) Guided Shift and (d) ZeroPur (Ours) reverse this deviation to varying degrees. (e) Feature trajectories under adversarial attack and the blur operation, indicating that simple blurring operator can provide a promising direction toward the natural image manifold.}
\label{motivation}
\end{figure*}

\subsection{Adversarial Purification in the Natural Image Manifold}
\label{subsec:adversarial_purification}

 Following the natural image manifold hypothesis~\cite{ma2018characterizing},~\cite{song2017pixeldefend}, natural images are assumed to reside on a specific manifold known as the natural image manifold. Given the classification loss function $\ell$, the learning process of DNNs finding optimal parameters $\theta$ can be formulated by:
\begin{equation}
\label{tr}
	\min_{\theta}\ell(f \circ g(\bm{x}), y; \theta),
\end{equation}
where a natural image $\bm{x}$ is embedded by an embedding function $f(\cdot) \in \mathbb{R}^{d}$ and then assigned its predicted label by a decision function $g(\cdot)$, and $y \in \mathbb{R}$ is its true label. Since the embedding space in which $f(\bm{x})$ is located can be considered as the common space of natural images associated with the decision made by $g$, this space can be viewed as an approximation of the natural image manifold $\mathcal{M}$. Therefore, this learning process can be regarded as an attempt to model the natural image manifold.

Adversarial attack crafts adversarial examples by optimizing the following objective:
\begin{equation}
\label{adv-attack}
	\max_{\Vert \bm{\delta}^{*} \Vert \leq \epsilon} \ell(f \circ g(\bm{x}+\bm{\delta}^{*}), y),
\end{equation}
where $\epsilon$ is the maximal norm which defines the set of allowed perturbations for a given example $\bm{x}$, in most contexts, such as the Projected Gradient Descent (PGD)~\cite{madry2017towards}, $\bm{\delta}^{*}$ is approximated by the local worst-cast $\bm{\delta}$. 

We are starting with formulating the forward process of $f$. Suppose $f: \mathbb{R}^{n} \rightarrow \mathbb{R}^{d}$ is twice differentiable, then the forward of $f$ can be defined as:
\begin{equation}
	f_{1...l}(\bm{x}) = f_{1} \circ f_{2} \circ ... \circ f_{l}(\bm{x}) = f_{l}(f_{l-1}(f_{2}...f_{1}(\bm{x}))),
\end{equation}
where $f_{i}$ typically refers to the output of the $\textit{i}^{th}$ layer in the function $f$. Based on the Taylor series, we have the following theorem:

\begin{theorem}
\label{error}
	Suppose $f: \mathbb{R}^{n} \rightarrow \mathbb{R}^{d}$ is twice differentiable at point $\bm{x}$. Let $e_{1} = \nabla f_{1}(\bm{x})^{T}\bm{\delta} + \frac{1}{2}\bm{\delta}^{T}\nabla^{2}f_{1}(\bar{\bm{x}})\bm{\delta}$, and there exists $w_{1} \in [0, 1]$ such that $\bar{\bm{x}} = w_{1}\bm{x} + (1 - w_{1})(\bm{x}+\bm{\delta})$, then we have the forward of $f$:
    \begin{align}
        f_{1...l}&(\bm{x}+\bm{\delta}) = f_{1...l}(\bm{x}) + e_{l}(\bm{\delta}),
    \end{align}
    where
    \begin{equation}
        \begin{aligned}
         \ \ \ e_{l}(\bm{\delta}) &= \nabla f_{l}(f_{1...l-1}(\bm{x}))^{T} e_{l-1}(\bm{\delta}) \\
        &\quad + \frac{1}{2} e_{l-1}(\bm{\delta})^{T} \nabla^{2} \bar{f}_{1...l-1}(\bm{x}) e_{l-1}(\bm{\delta}),
        \end{aligned}
    \end{equation}
and there exists $w_{l} \in [0, 1]$ such that $\bar{f}_{1...l-1}(\bm{x}) = w_{l}f_{1...l-1}(\bm{x}) + (1 - w_{l})(f_{1...l-1}(\bm{x}) + e_{l-1}(\bm{\delta}))$.
\end{theorem} 

\begin{proof}
	By a first-order Taylor series, we have:
	\begin{equation}
		f(\bm{x}) = f(\bm{x}_{0}) + \nabla f(\bm{x}_{0})^{T}(\bm{x} - \bm{x}_{0}) + \frac{1}{2}(\bm{x}-\bm{x}_{0})^{T}\nabla^{2}f(\bar{\bm{x}})(\bm{x}-\bm{x}_{0}).
	\end{equation}
	There exists $w \in [0, 1]$ such that $\bar{\bm{x}}=w\bm{x} + (1-w)\bm{x}_{0}$. For the forward of an adversarial example $\bm{x} + \bm{\delta}$ in the first layer $f_{1}$, we have:
	\begin{equation}
	\label{a-f1}
		f_{1}(\bm{x} + \bm{\delta}) = f_{1}(\bm{x}) + \underbrace{ {\nabla f_{1}(\bm{x})^{T}\bm{\delta} + \frac{1}{2}\delta^{T}\nabla^{2}f_{1}(\bar{\bm{x}})\bm{\delta}} }_{e_{1}(\bm{\delta})}.
	\end{equation}
	Plugging $f_{2}$ into \equref{a-f1}, we also have:
	\begin{align}
		& f_{1,2}(\bm{x} + \bm{\delta}) = f_{2}(f_{1}(\bm{x} + \bm{\delta})) = f_{2}(f_{1}(\bm{x}) + e_{1}(\bm{\delta}))\\
		&= \underbrace{f_{2}(f_{1}(\bm{x}))}_{f_{1,2}(\bm{x})} + \underbrace{\nabla f_{2}(f_{1}(\bm{x}))^{T}e_{1}(\bm{\delta}) + \frac{1}{2}e_{1}(\bm{\delta})^{T}\nabla^{2}\bar{f}_{1}(\bm{x})e_{1}(\delta)}_{e_{2}(\bm{\delta})}.
	\end{align}
	when $\bm{x} + \bm{\delta}$ is forward through the $l$-th layer, we have Theorem~\ref{error}.
\end{proof}

For notational simplicity, we denote $f_{1...l}(\bm{x}) = f(\bm{x})$ and $e_{l}(\bm{\delta}) = e(\bm{\delta})$ when $l$ is the last layer of $f$. Theorem~\ref{error} describes how the perturbation $\bm{\delta}$ is amplified layer by layer during the forward of $f$ so that the embedding of natural image $\bm{x}$ deviates manifold $\mathcal{M}$ and become the outlier, thereby demonstrates the reason for the incorrect decision made by $g$:
\begin{equation}
	g(f(\bm{x}+\bm{\delta})) = g(f(\bm{x}) + e(\bm{\delta})) \neq g(f(\bm{x})),
\end{equation}
where $e(\bm{\delta})$ can be viewed as the deviation distance from the natural image manifold. As shown in \figurename~\ref{motivation}\textcolor{blue}{(a)}, the adversarial images induced by malicious attacks on natural samples seriously deviate from the natural image manifold.

By this formulation, adversarial purification can be naturally seen as the process of estimating $e({\bm{\delta}})$ in the embedding space, the approximation of the natural image manifold. This means the adversarial embedding $f(\bm{x}_{\mathrm{adv}})=f(\bm{x}+\bm{\delta})$ can return the manifold along the same path $-e(\bm{\delta})$. While precise estimation of $e(\bm{\delta})$ is challenging, it is fortunate that only an approximation, denoted as $\tilde{e}(\bm{\delta})$, is sufficient to move the adversarial embedding back within the manifold. As can be observed in \figurename~\ref{motivation}\textcolor{blue}{(e)}, although the adversarial image after the blur operation does not fall exactly onto the original image, this favorable direction is sufficient to serve as a proxy target to guide the adversarial image correctly onto the natural manifold, as shown in \figurename~\ref{motivation}\textcolor{blue}{(c)} and \figurename~\ref{motivation}\textcolor{blue}{(d)}. We will elaborate on this novel purification pipeline ZeroPur in the following two subsections.

\subsection{Guided Shift}
\label{sec:gs}
Considering adversarial purification as an inverse process to adversarial attacks, we can emulate \equref{adv-attack} to formulate the optimization objective used to estimate $\tilde{e}(\bm{\delta})$:

\begin{align}
	\min_{\bm{\delta}_{\mathrm{pfy}}} \ell(f\  \circ \ &g(\bm{x}_{\mathrm{adv}} + \bm{\delta}_{\mathrm{pfy}}), y) \\
	\mathrm{s.t.}\ \ \ \Vert \bm{\delta}_{\mathrm{pfy}} \Vert \leq \epsilon_{\mathrm{pfy}}\ &\mathrm{and}\ e(\bm{\delta}_{\mathrm{pfy}})=-\tilde{e}(\bm{\delta}) \approx -e(\bm{\delta}),
\end{align}
where $\bm{\delta}_{\mathrm{pfy}}$ and $\epsilon_{\mathrm{pfy}}$ are defined to correspond to $\bm{\delta}$ and $\epsilon$ in \equref{adv-attack} to offset the perturbation. However, even if $\epsilon_{\mathrm{adv}}$ can be considered as a hyperparameter, the ground-truth label $y$ is not accessible. External model-based adversarial purification~\cite{yoon2021adversarial},~\cite{nie2022diffusion},~\cite{lee2023robust},~\cite{song2024mimicdiffusion} typically train a purification model to minimize the global $\ell_{\mathrm{pfy}}$. Auxiliary function-based adversarial method~\cite{mao2021adversarial},~\cite{shi2021online},~\cite{hwang2023aid} tends to design a suitable $\ell_{\mathrm{pfy}}$ to complete purification without relying on external generative models. They need to retrain the classifier or parameterized auxiliary function to effectively cooperate in removing adversarial perturbations, which means these methods all introduce parameters $\Theta$ in $\ell_{\mathrm{pfy}}$:
\begin{equation}
\min_{\bm{\delta}_{\mathrm{pfy}}} \ell_{\mathrm{pfy}}(f(\bm{x}_{\mathrm{adv}} + \bm{\delta}_{\mathrm{pfy}}); \Theta).\\
\end{equation}
Based on our assumption in  Section \ref{subsec:adversarial_purification} that adversarial images are outliers of the natural image manifold and the purification process can be considered as returning them to this manifold, we aim to design $\ell_{\mathrm{pfy}}$ without $\Theta$ by allowing them to  shift towards the manifold adaptively:
\begin{equation}
\min_{\bm{\delta}_{\mathrm{pfy}}} \ell^{\star}_{\mathrm{pfy}}(f(\bm{x}_{\mathrm{adv}} +\bm{\delta}_{\mathrm{pfy}}).
\end{equation}

\begin{algorithm}[!t]
\renewcommand{\algorithmicrequire}{\textbf{Input:}}
\renewcommand{\algorithmicensure}{\textbf{Output:}}
\caption{Guided Shift (GS)}
\label{gs}
\begin{algorithmic}[1]
\REQUIRE Adversarial example $\bm{x}_{\mathrm{adv}}$; Iterations $T_{\mathrm{g}}$; Step size $\eta_{1}$; Classifier $f$; Purification bound $\epsilon_{\mathrm{pfy}}$; Blurring operator $\mathrm{blur}(\cdot)$.
\ENSURE Guided result $\bm{x}_{\mathrm{g}}$
\STATE Random start $\bm{x}_{\mathrm{g}}^{0} \leftarrow \bm{x}_{\mathrm{adv}} + \varepsilon$ \hfill \textcolor{gray}{\small \textsf{// Noise initialization}}\\
\FOR {$t=0,1,2,...,T_{\mathrm{g}}-1$}
    \STATE $\bm{\tilde{x}}_{\mathrm{g}}^{t} \leftarrow \mathrm{blur}(\bm{x}_{\mathrm{g}}^{t})$ \hfill \textcolor{gray}{\small \textsf{// Generate manifold guidance}}
    \STATE $\bm{z} \leftarrow f(\bm{x}_{\mathrm{g}}^{t}), \bm{\tilde{z}} \leftarrow f(\bm{\tilde{x}}_{\mathrm{g}}^{t})$  \hfill \textcolor{gray}{\small \textsf{// Extract embeddings}}
    \STATE $\bm{x}_{\mathrm{g}}^{t+1} \leftarrow \bm{x}_{\mathrm{g}}^{t} + \eta_{1} \cdot \mathrm{sgn}\bigl(\nabla_{\bm{x}_{\mathrm{g}}^{t}}d(\bm{z},\bm{\tilde{z}})\bigr)$ \hfill \textcolor{gray}{\small \textsf{// Purify via Eq.\eqref{eq:guided_shift}}}
    \STATE $\bm{x}_{\mathrm{g}}^{t+1} \leftarrow \mathrm{Clip} (\bm{x}^{t+1}_{\mathrm{g}}, -\epsilon_{\mathrm{pfy}}, \epsilon_{\mathrm{pfy}})$ 
    \hfill \textcolor{gray}{\small \textsf{// Constrain perturbation}}
    \STATE $\bm{x}_{\mathrm{g}}^{t+1} \leftarrow \mathrm{Clip}(\bm{x}^{t+1}_{\mathrm{g}}, 0, 1)$
    \hfill \textcolor{gray}{\small \textsf{// Project to pixel space}}
\ENDFOR
\STATE $\bm{x}_{\mathrm{g}} \leftarrow \bm{x}_{\mathrm{g}}^{T_{\mathrm{g}}}$
\RETURN $\bm{x}_{\mathrm{g}}$
\end{algorithmic}
\end{algorithm}

\begin{figure}[!t]
    \centering
    \subfloat{%
        \includegraphics[width=0.48\linewidth]{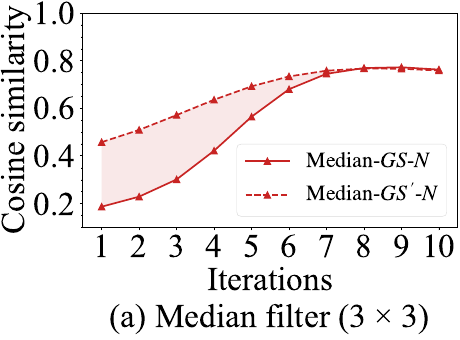}
        \label{fig:median-3}
    }
    \hfill
    \subfloat{%
        \includegraphics[width=0.48\linewidth]{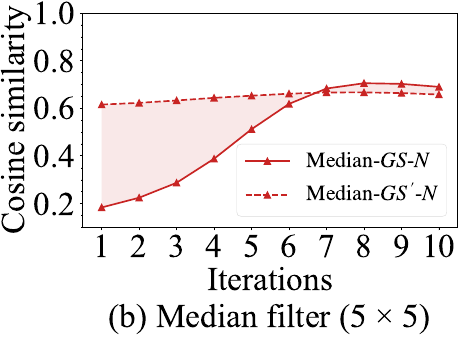}
        \label{fig:median-5}
    }\\
    \subfloat{%
        \includegraphics[width=0.48\linewidth]{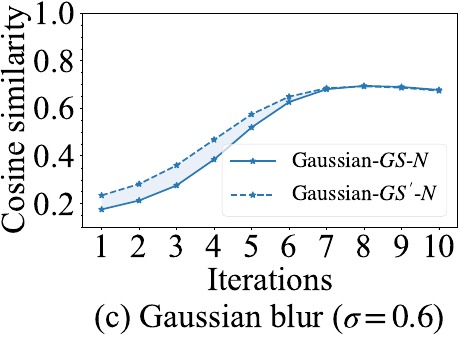}
        \label{fig:gaussian-0.6}
    }
    \hfill
    \subfloat{%
        \includegraphics[width=0.48\linewidth]{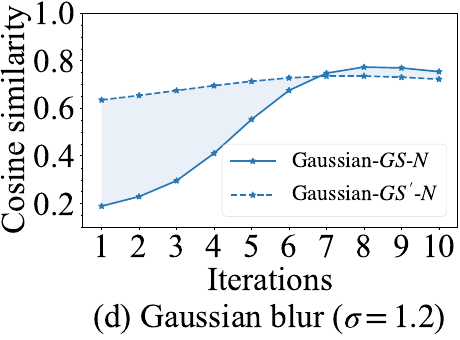}
        \label{fig:gaussian-1.2}
    }
    \caption{Details of GS on CIFAR-10.}
    \label{gs-limit}
\end{figure}

We are starting to investigate whether a simple image transformation, such as \textit{color jitter}, \textit{grayscale}, \textit{gaussian blur}, \textit{solarization}, and \textit{equalization}, can shift adversarial images towards the natural image manifold. Fortunately, as shown in \figurename~\ref{motivation}\textcolor{blue}{(e)}, the blurring operation nearly provides a genuine direction toward the natural samples for the adversarial examples. This phenomenon suggests that a simple blurring operator can bring adversarial examples closer to the manifold. 

Therefore, we can accumulate this reversion to guide the adversarial examples back towards the manifold. For instance, we can pull the distance between adversarial examples and their blurred counterparts in the embedding space, which allows adversarial examples to converge closer to the natural image manifold. The distance of feature embeddings is defined by the Cosine Similarity:
\begin{equation}
\label{cosine}
	d(\bm{z}_{\mathrm{adv}}, \bm{z}_{\mathrm{adv}}^{\prime}) = \frac{\bm{z}_{\mathrm{adv}}\cdot \bm{z}_{\mathrm{adv}}^{\prime}}{\Vert \bm{z}_{\mathrm{adv}} \Vert \Vert \bm{z}_{\mathrm{adv}}^{\prime} \Vert},
\end{equation}
where $\bm{z}_{\mathrm{adv}}$ and $\bm{z}_{\mathrm{adv}}^{\prime}$ are the embeddings of adversarial examples and their blurred counterparts, respectively.  we can shift the adversarial example $\bm{x}_{\mathrm{adv}}$ by the gradient of \equref{cosine}:
\begin{equation}
\label{step}
	\bm{x}_{\mathrm{adv}}^{+} =\bm{x}_{\mathrm{adv}} + \eta_{1}\cdot \nabla_{\bm{x}_{\mathrm{adv}}}d(f(\bm{x}_{\mathrm{adv}}), f(\bm{x}_{\mathrm{adv}}^{\prime})).
\end{equation}

Controlling the magnitude of the blurring applied carefully is essential to prevent excessive blurring that could render adversarial examples unrecognizable by the classifier. However, a small magnitude is not sufficient to shift images to reverse the distortion caused by adversarial perturbations. We therefore consider to move $\bm{x}_{\mathrm{adv}}$ by \equref{step} with a small step size $\eta_{1}$ iteratively. This process is referred to as \textbf{G}uided \textbf{S}hift (GS). In each step of GS, we apply the same update rule:
\begin{equation}
    \bm{x}_{\mathrm{g}}^{t+1} = \bm{x}_{\mathrm{g}}^{t} + \eta_{1} \cdot \nabla_{\bm{x}_{\mathrm{g}}^{t}} d(f(\bm{x}_{\mathrm{g}}^{t}), f(\bm{\tilde{x}}_{\mathrm{g}}^{t}))
    \label{eq:guided_shift}
\end{equation}
\begin{equation}
    \bm{\tilde{x}}_{\mathrm{g}}^{t+1} = \mathrm{blur}(\bm{x}_{\mathrm{g}}^{t+1})
    \label{eq:blur_update}
\end{equation}
where $blur(\cdot)$ is the blurring operator and $\bm{x}_{\mathrm{g}}^{0} := \bm{x}_{\mathrm{adv}}$. The workflow of GS is shown in Algorithm~\ref{gs}, where we use $\mathrm{sgn}(\cdot)$ to regulate the step size.

To demonstrate the process of GS, we employ two blurring operators: the Median filter and Gaussian blur to compute the cosine similarity between single-step result $\bm{x}_{\mathrm{g}}^{t}$ of GS and the natural image $\bm{x}$ ($\star$-$GS$-$N$), as well as between the blurred counterpart $\bm{\tilde{x}}_{\mathrm{g}}^{t}$ of $\bm{x}_{\mathrm{g}}^{t}$ and $\bm{x}$ ($\star$-$GS^{\prime}$-$N$), as shown in \figurename~\ref{gs-limit}. Interestingly, as GS progresses, both cosine similarities consistently increase, overcoming the inherent limitations of a single blur. Its effectiveness can be qualitatively described by \figurename~\ref{motivation}\textcolor{blue}{(c)}, where adversarial examples, guided by their blurred counterparts, gradually approach the natural manifold.

 \subsection{Adaptive Projection}
 
We can also observe an interesting phenomenon in Fig.~\ref{gs-limit}. Regardless of the type of blurring operators, dash lines ($\star$-$GS^{\prime}$-$N$) and solid lines ($\star$-$GS$-$N$) tend to converge. Therefore, we can have the following assumption: 
\begin{assumption}
There is a limitation for Guided Shift in that the adversarial images can be guided by their blurred counterparts only toward the manifold rather than fully back to the manifold. In other words, this indicates that Guided Shift approximates a convex function.
\end{assumption}

\begin{figure}[!t]
    \centering
    \subfloat[AP w/o. PR in Adv $\bm{x}_{\mathrm{init}}$]{%
        \label{fig:ap-adv}
        \includegraphics[width=0.46\linewidth]{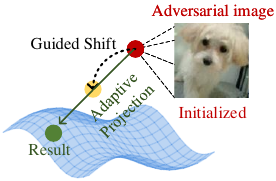}
    }
    \hfill
    \subfloat[AP in Adv $\bm{x}_{\mathrm{init}}$]{%
        \label{ap-adv-reg}
        \includegraphics[width=0.46\linewidth]{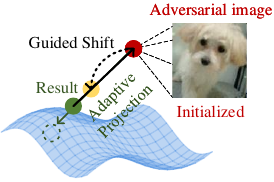}
    }\\
    \subfloat[AP w/o. PR in Nat $\bm{x}_{\mathrm{init}}$]{%
        \label{ap-nat}
        \includegraphics[width=0.46\linewidth]{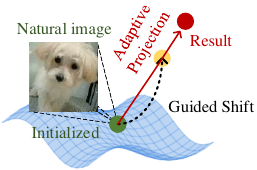}
    }
    \hfill
    \subfloat[AP in Nat $\bm{x}_{\mathrm{init}}$]{%
        \label{ap-nat-reg}
        \includegraphics[width=0.46\linewidth]{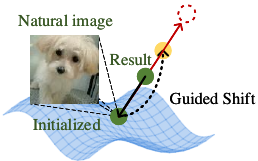}
    }

    \caption{Intuitive explanation of AP in the context of adversarial or natural $\bm{x}_{\mathrm{init}}$. PR: the perceptual regularization (\equref{F2}).}
    \label{fig:toy}
\end{figure}

\begin{algorithm}[!t]
\renewcommand{\algorithmicrequire}{\textbf{Input:}}
\renewcommand{\algorithmicensure}{\textbf{Output:}}
\caption{Adaptive Projection (AP)}
\label{alg:ap}
\begin{algorithmic}[1]
\REQUIRE Input image $\bm{x}_{\mathrm{init}}$; Guided result $\bm{x}_{\mathrm{g}}$; Iterations $T_{\mathrm{p}}$; Classifier $f$; Purification bound $\epsilon_{\mathrm{pfy}}$; Candidate layers $S$.
\ENSURE Adaptive result $\bm{x}_{\mathrm{p}}$
\STATE Initialize $\bm{x}_{\mathrm{p}}^{0} \leftarrow \bm{x}_{\mathrm{init}}$
\STATE Set step size $\eta_{2} \leftarrow \epsilon_{\mathrm{pfy}} / T_{\mathrm{p}}$
\FOR {$t=0,1,2,...,T_{\mathrm{p}}-1$}
    \STATE $\mathcal{L} \leftarrow 0$ \hfill \textcolor{gray}{\small \textsf{// Initialize loss for current step}}
    \FOR {$l \in S$}
        \STATE $\Delta\bm{u}_{\mathrm{g}}^{l} \leftarrow f_{l}(\bm{x}_{\mathrm{g}}) - f_{l}(\bm{x}_{\mathrm{init}})$ \hfill \textcolor{gray}{\small \textsf{// GS direction}}
        \STATE $\Delta\bm{u}_{\mathrm{p}}^{l} \leftarrow f_{l}(\bm{x}_{\mathrm{p}}^{t}) - f_{l}(\bm{x}_{\mathrm{init}})$ \hfill \textcolor{gray}{\small \textsf{// Current projection}}
        \STATE $\mathcal{L} \leftarrow \mathcal{L} + (-\Delta\bm{u}_{\mathrm{g}}^{l}\cdot\Delta\bm{u}_{\mathrm{p}}^{l})$ \hfill \textcolor{gray}{\small \textsf{// Accumulate loss}}
    \ENDFOR
    \STATE $\mathcal{L} \leftarrow \frac{\lambda_{1}}{\Vert S \Vert}\mathcal{L} + \lambda_{2}\Vert \phi(\bm{x}_{\mathrm{p}}^{t}) - \phi(\bm{x}_{\mathrm{init}}) \Vert_{2}$ \hfill \textcolor{gray}{\small \textsf{// Eq.\eqref{AP} with regularization}}
    \STATE $\bm{x}_{\mathrm{p}}^{t+1} \leftarrow \bm{x}_{\mathrm{p}}^{t} - \eta_{2} \cdot \mathrm{sgn}(\nabla_{\bm{x}_{\mathrm{p}}^{t}}\mathcal{L})$ \hfill \textcolor{gray}{\small \textsf{// Update projection}}
    \STATE $\bm{x}_{\mathrm{p}}^{t+1} \leftarrow \mathrm{Clip}(\bm{x}_{\mathrm{p}}^{t+1}, -\epsilon_{\mathrm{pfy}},\epsilon_{\mathrm{pfy}})$ \hfill \textcolor{gray}{\small \textsf{// Constrain perturbation}}
    \STATE $\bm{x}_{\mathrm{p}}^{t+1} \leftarrow \mathrm{Clip}(\bm{x}_{\mathrm{p}}^{t+1}, 0, 1)$ \hfill \textcolor{gray}{\small \textsf{// Project to pixel space}}
\ENDFOR
\STATE $\bm{x}_{\mathrm{p}} \leftarrow \bm{x}_{\mathrm{p}}^{T_{\mathrm{p}}}$
\RETURN $\bm{x}_{\mathrm{p}}$
\end{algorithmic}
\end{algorithm}

Certainly, this limitation arises from the exceedingly blurred images whose embeddings cannot be recognized correctly by the classifier. To fully return adversarial images to the manifold, we must overcome this limitation. A natural intuition is to allow current images to move adaptively instead of being guided by blurred counterparts, which requires us to target a basic direction for current images. Fortunately, the result of Guided Shift already provides this basic direction. Therefore, we propose Adaptive Projection (AP) that can be defined as:
\begin{align}
\label{AP}
	\min_{\bm{x}_{\mathrm{p}}}\lambda_{1}\frac{1}{\Vert S \Vert}\sum_{l \in S}\mathcal{L}_{l}(\bm{x}_{\mathrm{init}}, \bm{x}_{\mathrm{g}}, \bm{x}_{\mathrm{p}})& +\lambda_{2} \Vert \phi(\bm{x}_{\mathrm{p}}) - \phi(\bm{x}_{\mathrm{init}}) \Vert_{2} \notag \\ 
    \mathrm{s.t.}\ \ \ \Vert \bm{x}_{\mathrm{p}} - \bm{x}_{\mathrm{init}} \Vert& \leq \bm{\delta}_{\mathrm{pfy}},
\end{align}
where $S \subseteq L$ as a candidate in the set $L = \{ l_{1},l_{2},...,l_{m} \}$ of a $m$ layers model $f$,  $\Vert S \Vert$ denotes the number of elements of the set $S$. $\bm{x}_{\mathrm{init}}$ represents the input image (i.e., $\bm{x}_{\mathrm{init}}=\bm{x}_{\mathrm{adv}}$ when the input image is adversarial). For the $l^{th}$ layer of classifiers, the first term of \equref{AP} can be written as: 
\begin{equation}
 \label{ila}
 	\mathcal{L}_{l}(\bm{x}_{\mathrm{init}}, \bm{x}_{
 	\mathrm{g}}, \bm{x}_{\mathrm{p}}) = -\Delta\bm{u}^{l}_{\mathrm{p}} \cdot \Delta\bm{u}^{l}_{\mathrm{g}},
 \end{equation}
 where $\bm{x}_{\mathrm{p}}$ is the final result, $\Delta\bm{u}^{l}_{\mathrm{p}}$ and $\Delta\bm{u}^{l}_{\mathrm{g}}$ are two vectors of flattened feature maps defined as follow:
 \begin{equation}
 	\Delta\bm{u}^{l}_{\mathrm{p}} = f_{l}(\bm{x}_{\mathrm{p}}) - f_{l}(\bm{x}_{\mathrm{init}}),\ \ \  \Delta\bm{u}^{l}_{\mathrm{g}} = f_{l}(\bm{x}_{\mathrm{g}}) - f_{l}(\bm{x}_{\mathrm{init}}),
 \end{equation}
where $f_{l}$ denotes feature maps at layer $l$ of the classifier, $\bm{x}_{\mathrm{p}}$ is initialized by $\bm{x}_{\mathrm{init}}$. Maximizing \equref{ila} is equivalent to maximizing the projection of $\bm{u}^{l}_{\mathrm{p}}$ onto $\bm{u}^{l}_{\mathrm{g}}$ since $\Vert \bm{u}^{l}_{\mathrm{g}} \Vert$ is a constant. The increase in projection allows $\bm{x}_{\mathrm{p}}$ to move freely along the direction of GS, enabling it to independently approach the natural image manifold.

However, we typically do not know whether an image is adversarial in real-world applications (i.e., $\bm{x}_{\mathrm{init}}=\bm{x}$ when $\bm{x}$ is natural), repeatedly guiding images by blurred counterparts may cause natural images to deviate from the manifold. As shown in \figurename~\ref{fig:toy}, this deviation will be exacerbated by AP. Therefore, we introduce a perceptual regularization term in \equref{AP}. Let $F_{1}$ is the dynamic of \equref{ila}:
\begin{equation}
\label{ap}
F_{1} = \lambda_{1}\cdot\nabla_{\bm{x}_{\mathrm{p}}} \frac{1}{\Vert S \Vert}\sum_{l \in S}\mathcal{L}_{l}(\bm{x}_{\mathrm{init}}, \bm{x}_{\mathrm{g}}, \bm{x}_{\mathrm{p}})\ ,
\end{equation}
if there exists a momentum $F_{2}$, such that:
	\begin{equation}
	\left\{
		\begin{array}{lll}
			F_{1} > F_{2}, & \bm{x}_{\mathrm{init}} \ is \ adversarial, \\
			F_{1} < F_{2}, & \bm{x}_{\mathrm{init}} \  is \ natural, \\
		\end{array}
	\right.
	\end{equation}
then maximizing \equref{ila} allows the result $\bm{x}_{p}$ to locate in the natural image manifold. In this work, the dynamic $F_{2}$ can be computed by LPIPS distance~\cite{zhang2018unreasonable}. The LPIPS distance $d(\bm{x}_{1}, \bm{x}_{2})$ between images $\bm{x}_{1}$ and $\bm{x}_{2}$ is then defined as $d(\bm{x}_{1}, \bm{x}_{2}) \triangleq \Vert \phi(\bm{x}_{1}) - \phi(\bm{x}_{2}) \Vert_{2}$. Let $\hat{f}(\bm{x})$ denote channel-normalized activations at the $l$-th layer of the classifier. Then, $\phi(\bm{x}) \triangleq (\frac{\hat{f}(\bm{x})}{\sqrt{w_{1}h_{1}}},...,\frac{\hat{f}(\bm{x})}{\sqrt{w_{m}h_{m}}})$, where $w_{l}$ and $h_{l}$ are the width and height of activations of layer $l$, respectively.

Finally, the momentum $F_{2}$ can be written as:
\begin{equation}
\label{F2}
	\centering
	F_{2} = -\lambda_{2}\cdot\nabla_{\bm{x}_{\mathrm{p}}}\Vert \phi(\bm{x}_{\mathrm{p}}) - \phi(\bm{x}_{\mathrm{init}}) \Vert_{2}.
\end{equation}
In contrast to the original LPIPS objective of approximating human perceptual judgments, our goal here is to ensure that the perceptual distance between $\bm{x}_{\mathrm{p}}$ and $\bm{x}_{\mathrm{init}}$, as measured by the classifier itself, remains small. The workflow of AP is shown in Algorithm~\ref{alg:ap}.

\section{Experiments}
\label{sec:experiments}
In this section, we evaluate the effectiveness of ZeroPur under various adversarial attacks across different datasets and model architectures, including both common attacks and adaptive attacks. We also conduct sensitivity analysis and ablation study to thoroughly examine the impact of ZeroPur’s components, and provide visualizations to illustrate how it guides the purification process.

\subsection{Experimental Settings}
\label{experiments}
\textbf{Datasets and Metrics.}
Three benchmarks CIFAR-10~\cite{krizhevsky2009learning}, CIFAR-100~\cite{krizhevsky2009learning}, and ImageNet-1K~\cite{deng2009imagenet} are considered to evaluate our method. In all experiments, we consider two metrics to evaluate the performance of all methods: standard accuracy (SA) and robust accuracy (RA). The standard accuracy measures the performance of the defense method on natural images, while the robust accuracy measures the performance on adversarial images.

\textbf{SOTA Methods.}
 We compare our method with the state-of-the-art adversarial training methods reported in the standard benchmark RobustBench~\cite{croce2021robustbench}, as well as with other representative adversarial training and purification methods~\cite{li2023data},~\cite{xu2023exploring},~\cite{zhounasty},~\cite{jin2025s2o},~\cite{chen2026data},~\cite{shi2021online},~\cite{mao2021adversarial},~\cite{kulkarni2024interpretability},\cite{yoon2021adversarial},~\cite{ughini2022trust},~\cite{nie2022diffusion},~\cite{lee2023robust},~\cite{lin2024adversarial},~\cite{bai2024diffusion},~\cite{chen2024data},~\cite{song2024mimicdiffusion},~\cite{pei2025diffusion}. In addition, Gaussian Blur is included as a simple preprocessing baseline to distinguish the effect of blur-based transformation from the proposed feature-guided purification mechanism in ZeroPur. The adversarial training methods provided by RobustBench~\cite{croce2021robustbench} are available at \href{https://robustbench.github.io}{https://robustbench.github.io}. The code for adversarial purification methods is obtained from their respective official implementations.

\textbf{Adversarial Attacks.}
We evaluate our methods with four attacks: PGD~\cite{madry2017towards}, AutoAttack~\cite{croce2020reliable}, DI$^{2}$-FGSM~\cite{xie2019improving}, and BPDA~\cite{athalye2018obfuscated}. PGD is a widely used iterative adversarial attack, often considered a strong first-order baseline. AutoAttack is a more powerful attack commonly used in most defense studies. We also use DI$^{2}$-FGSM to seek whether an attack robust to the blurring operator can affect our method. Additionally, we consider BPDA to attack our purification module in the worst-case scenario.

\begin{itemize}
\item \textbf{PGD}~\cite{madry2017towards}. The Projected Gradient Descent (PGD) attack generates adversarial examples by iteratively applying gradient ascent on the input and projecting the perturbed input back into the $\epsilon$-ball around the original image. In our experiments, we use a step size of $\alpha = 2/255$ and set the perturbation bound to $L_\infty = 8/255$.
\item \textbf{AutoAttack}~\cite{croce2020reliable}.\ \ We use AutoAttack\footnote{\url{https://github.com/fra31/auto-attack}} to compare with the start-of-the-art methods. There are two versions of AutoAttack: (i) the \textsc{Standard} including AGPD-CE, AGPD-T, FAB-T, and Square, and (ii) the \textsc{Rand} version including APGD-CE and APGD-DLR. Considering that most of the adversarial purifications choose the \textsc{Rand} version, all the performance in this work we report is also in the \textsc{Rand} version. 
\item \textbf{DI$^{2}$-FGSM}~\cite{xie2019improving}. DI$^{2}$-FGSM crafts adversarial examples by applying various transformations, enhancing their robustness against blurring operations. We use this attack to evaluate ZeroPur in Section~\ref{ab}, implemented by torchattacks\footnote{\url{https://github.com/Harry24k/adversarial-attacks-pytorch}} ~\cite{kim2020torchattacks}.
\item \textbf{BPDA}~\cite{athalye2018obfuscated}. We use BPDA, approximating the gradient of the purifier module (ZeroPur) as 1 during the backward pass. The 10, 20, and 40 iterations are applied in our experiment. Other settings are the same as those used in PGD attack.
\end{itemize}

\begin{table}[!t]
  \centering
  \caption{Natural accuracy of victim classifiers under different augmentation strategies: “Vanilla” (no augmentation), “Base” (ResizeCrop + HorizontalFlip), and “Strong” (ResizeCrop, ColorJitter, Grayscale, Solarization, Equalization, HorizontalFlip).}
\fontsize{8.5}{9}\selectfont 
\renewcommand{\tabcolsep}{5pt}
\renewcommand{\arraystretch}{1.0}
    \begin{tabular}{cccc}
    \toprule
    \multirow{2}{*}{Victim Classifier} & \multirow{2}{*}{Augmentation} & \multicolumn{2}{c}{Natural Accuracy (\%)} \\
    \cmidrule(lr){3-4}  
    &   & CIFAR-10 & CIFAR-100 \\
    \midrule 
    \multirow{3}{*}{ResNet-18} & Vanilla& 83.80 & 57.01 \\
    & Base &  \textbf{93.10} & \textbf{71.58} \\
    & Strong & 91.08 & 68.53 \\
    \midrule   
    \multirow{3}{*}{WRN-28-10} & Vanilla & 91.34 & 71.57 \\
    & Base &  \textbf{93.83}  & \textbf{74.59} \\
    & Strong &  91.10     & 67.65 \\
    \bottomrule
    \end{tabular}%
  \label{details}%
\end{table}%

\textbf{Victim Classifiers.}
The above adversary will attack three victim classifiers, including ResNet-18~\cite{he2016deep}, ResNet-50~\cite{he2016deep}, and WideResNet-28-10~\cite{zagoruyko2016wide}. To evaluate the defense performance of ZeroPur against victim classifiers trained with different data augmentation strategies (see Section~\ref{ab} for details), instead of using the pretrained models from PyTorch or the open source model library timm\footnote{\url{https://github.com/rwightman/pytorch-image-models}} as we do for ResNet-50, we implement the victim classifiers for ResNet-18 and WideResNet-28-10 ourselves using three data augmentation strategies: ‘Vanilla', ‘Base', and ‘Strong'. Table~\ref{details} presents the training settings and natural accuracies. ‘Base' is the most commonly used data augmentation strategy in adversarial training and purification methods for better natural accuracy.

All classifiers were trained with the SGD optimizer with a cosine decay learning rate schedule and a linear warm-up period of 10 epochs. The weight decay is $5.0 \times 10^{-4}$ and the momentum is $0.9$. The initial learning rate is set to $0.1$. Classifiers were trained for 120 epochs on 4 Tesla V100 GPUs, where the batch size is 512 per GPU for ResNet-18 and 128 per GPU for WideResNet-28-10.

It is worth noting that the training here is conducted solely to provide pretrained models under different training settings for a comprehensive evaluation of ZeroPur. In practice, ZeroPur can defend against any already trained off-the-shelf model.

\begin{figure}[tb]
    \centering
    \subfloat[Robust accuracy]{%
        \includegraphics[width=0.48\linewidth]{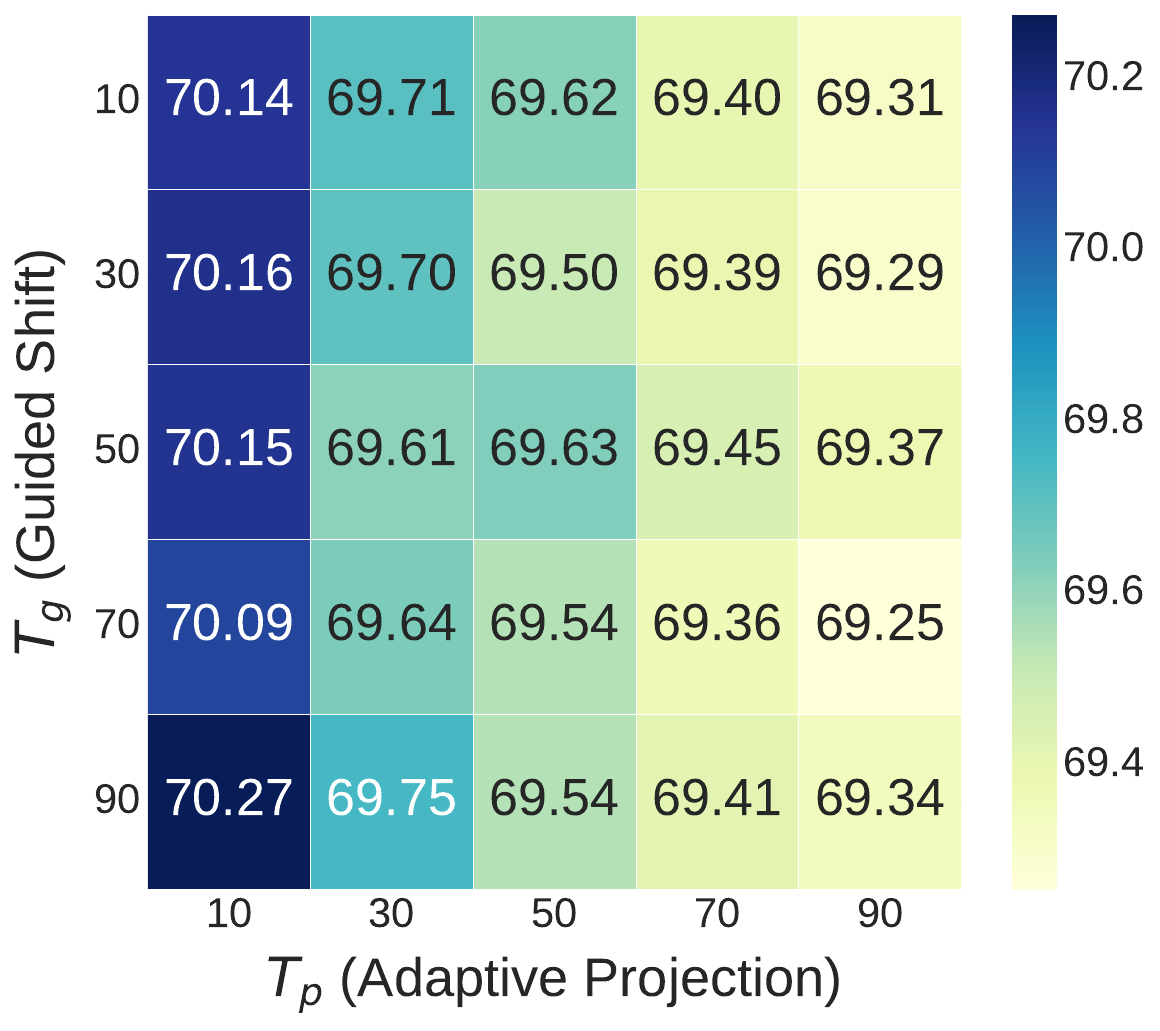}
    }
    \hfill
    \subfloat[Standard accuracy]{%
        \includegraphics[width=0.48\linewidth]{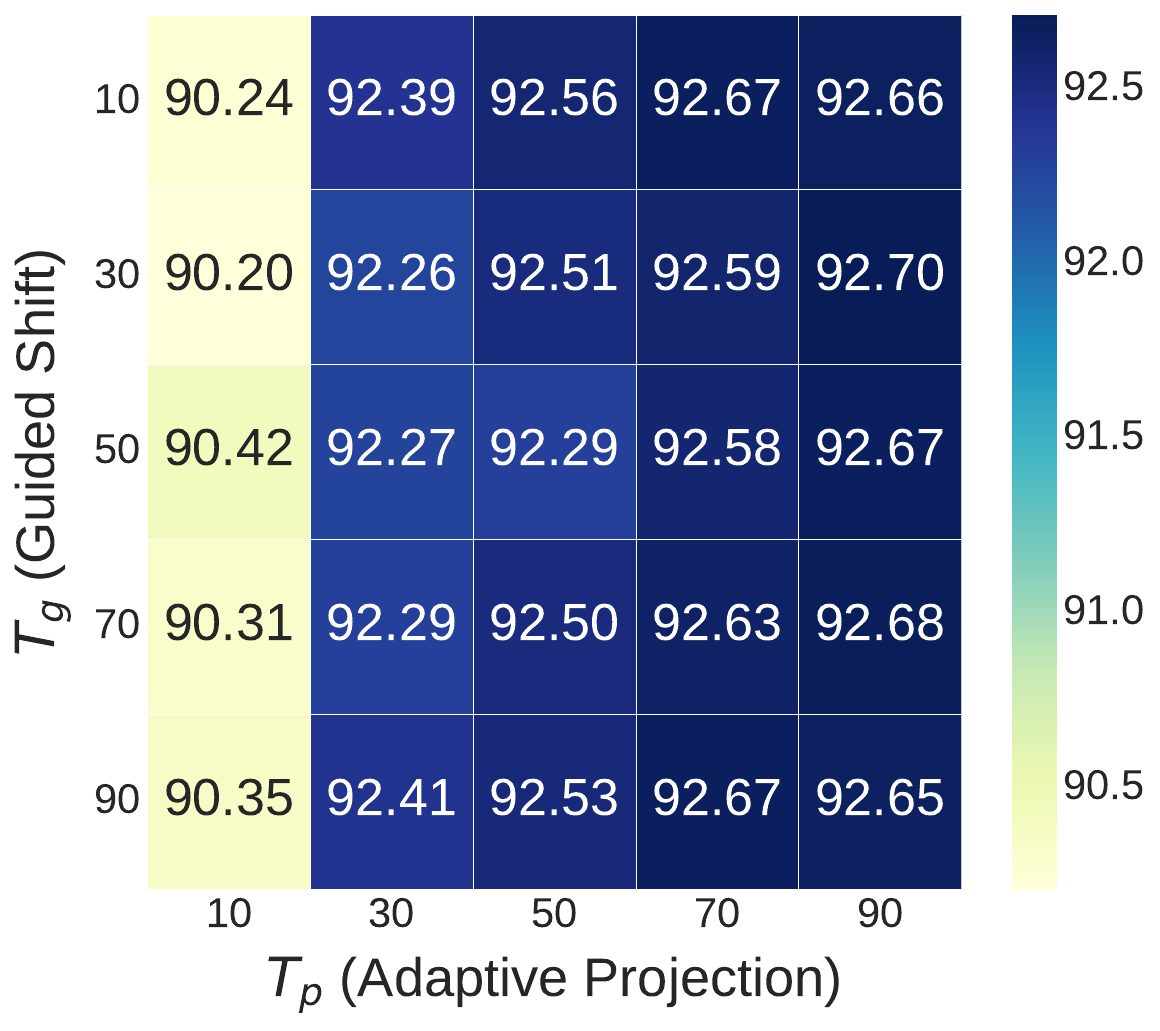}
    }
\caption{Robust and standard accuracy (\%) of ZeroPur on CIFAR-10 under different iteration numbers of Guided Shift ($T_{g}$) and Adaptive Projection ($T_{p}$). The performance remains stable across a wide range of $(T_{g}, T_{p})$ settings.}
\label{fig:parameter}
\end{figure}

\textbf{Implementation Details.} In the Adaptive Projection stage of ZeroPur, the adaptation strength to natural and adversarial samples is governed by two key hyperparameters, $\lambda_{1}$ and $\lambda_{2}$. Their optimal values were determined through a lightweight grid search. Specifically, for CIFAR-10, we adopt $(10^{-3},10^{1})$ and $(10^{-3},10^{2})$ on ResNet-18 and WRN-28-10, respectively; for CIFAR-100, we use $(10^{-4},1)$ and $(2\times10^{-4},10^{1})$ on ResNet-18 and WRN-28-10, respectively; and for ImageNet with ResNet-50, we set $(1.5 \times 10^{-5},1)$. Furthermore, the effect of the iteration numbers $T_{g}$ (Guided Shift) and $T_{p}$ (Adaptive Projection) was analyzed on CIFAR-10 using heatmaps of robust and standard accuracy. These analyses demonstrate the inherent stability of ZeroPur with respect to these iteration parameters, as illustrated in Fig.~\ref{fig:parameter}. Consequently, and without loss of generality, $T_{g}$ is uniformly set to $10$ and $T_{p}$ to $50$ across all subsequent experiments. Finally, the purification radius $\epsilon_{pfy}$ is empirically set to $1.25 \times \epsilon_{atk}$ for the effective elimination of adversarial perturbations.

\begin{table*}[tb]
\caption{Performance comparison with adversarial training (AT) and auxiliary-based purification (ABP) methods under AutoAttack $\ell_\infty$ ($\epsilon = 8/255$). \textcolor{green}{$\bf\diamond$} and \textcolor{blue}{$\bf\diamond$} denote AT and ABP methods, respectively. $^{\ddagger}$ indicates the requirement of extra data. * indicates the method uses WideResNet-34-10 as a classifier.}
\centering
\fontsize{8.5}{9}\selectfont 
\renewcommand{\tabcolsep}{5pt}
\renewcommand{\arraystretch}{1.0}
\begin{tabular}{cclcclcc} 
    \toprule
    \multicolumn{2}{c}{Require Training}  
    & \multicolumn{3}{c}{ResNet-18}  
    & \multicolumn{3}{c}{WideResNet-28-10} \\ 
    \cmidrule(lr){1-2} \cmidrule(lr){3-5} \cmidrule(lr){6-8} 
    Cls & Aux & Method & SA(\%) & RA(\%) & Method & SA(\%) & RA(\%) \\
    
    \midrule
    \rowcolor{gray!10} \multicolumn{8}{c}{CIFAR-10} \\ 
    \midrule
    
    \checkmark & \xmark & \textcolor{green}{$\bf\diamond$} Gowal et al.~\cite{gowal2021improving}$^{\ddagger}$ & 87.35 & 58.50 & 
    \textcolor{green}{$\bf\diamond$} Rebuffi et al.~\cite{rebuffi2021data}$^{\ddagger}$ &  85.97 &  60.73 \\
    \checkmark & \xmark & \textcolor{green}{$\bf\diamond$} Sehwag et al.~\cite{sehwag2021robust}$^{\ddagger}$ & 84.60 & 55.70 & 
    \textcolor{green}{$\bf\diamond$} Gowal et al.~\cite{gowal2021improving}$^{\ddagger}$ & 87.50 & 63.38 \\
    \checkmark & \xmark & \textcolor{green}{$\bf\diamond$} Rade et al.~\cite{rade2021helper}$^{\ddagger}$ & 86.86 & 57.09 & 
    \textcolor{green}{$\bf\diamond$} Pang et al.~\cite{pang2022robustness}$^{\ddagger}$ & 88.61 & 61.04 \\
    \checkmark & \xmark & \textcolor{green}{$\bf\diamond$} Addepalli et al.~\cite{addepalli2022efficient} & 85.71 & 52.50 &
    \textcolor{green}{$\bf\diamond$} Xu et al.~\cite{xu2023exploring}$^{\ddagger}$ & \textbf{93.69} & 63.89\\
    \checkmark & \xmark & \textcolor{green}{$\bf\diamond$} Li et al.~\cite{li2023data} & 83.45 & 52.52 & 
    \textcolor{green}{$\bf\diamond$} Jin et al.~\cite{jin2025s2o}$^{\ddagger}$ & 88.93 & 61.45 \\
   
    \checkmark & \xmark & \textcolor{green}{$\bf\diamond$} Zhou et al.~\cite{zhounasty} & 90.86 & 50.18 & 
    \textcolor{green}{$\bf\diamond$} Chen et al.~\cite{chen2026data}  & 89.50  & 62.76  \\

    \cdashline{1-8} 
    
    \checkmark & \xmark & \textcolor{blue}{$\bf\diamond$} Shi et al.~\cite{shi2021online} & 84.07 & 66.62 & \textcolor{blue}{$\bf\diamond$} Shi et al.~\cite{shi2021online} & 91.89 & 68.56 \\
    \xmark & \checkmark & \textcolor{blue}{$\bf\diamond$} Mao et al.~\cite{mao2021adversarial} & -- & 58.20 & \textcolor{blue}{$\bf\diamond$} Mao et al.~\cite{mao2021adversarial} & -- & 67.15 \\
    \xmark & \checkmark & \textcolor{blue}{$\bf\diamond$} Kulkarni et al.~\cite{kulkarni2024interpretability} & 85.18 & 54.53 & \textcolor{blue}{$\bf\diamond$} Hwang et al.~\cite{hwang2023aid}* & 82.22 & 56.63 \\
    
    {\color{lightgray}\xmark} & {\color{lightgray}\xmark} & {\color{lightgray}\hphantom{\textcolor{blue}{$\bf\diamond$}} Gaussian Blur} & {\color{lightgray}76.35} & {\color{lightgray}29.93} & {\color{lightgray}\hphantom{\textcolor{blue}{$\bf\diamond$}} Gaussian Blur} & {\color{lightgray}76.24} & {\color{lightgray}28.16} \\
    
    \xmark & \xmark & \textcolor{blue}{$\bf\diamond$} \textbf{GS} & 49.56 & 58.82 & \textcolor{blue}{$\bf\diamond$} \textbf{GS} & 50.24 & 59.68 \\
    \cellcolor{my_color}\xmark & \cellcolor{my_color}\xmark & \cellcolor{my_color}\textcolor{blue}{$\bf\diamond$} \textbf{GS+AP (ZeroPur)} & \cellcolor{my_color}\textbf{92.56} & \cellcolor{my_color}\textbf{69.62} & \cellcolor{my_color}\textcolor{blue}{$\bf\diamond$} \textbf{GS+AP (ZeroPur)} & \cellcolor{my_color}91.81 & \cellcolor{my_color}\textbf{68.60} \\
    
    \midrule
    \rowcolor{gray!10} \multicolumn{8}{c}{CIFAR-100} \\
    \midrule
    
    \checkmark & \xmark & \textcolor{green}{$\bf\diamond$} Rade et al.~\cite{rade2021helper}$^{\ddagger}$ & 61.50 & 28.88 & 
    \textcolor{green}{$\bf\diamond$} Rebuffi et al.~\cite{rebuffi2021data}$^{\ddagger}$ &  59.18 &  30.81 \\
    \checkmark & \xmark & \textcolor{green}{$\bf\diamond$} Addepalli et al.~\cite{addepalli2022efficient} & \textbf{65.45} & 27.69 & \textcolor{green}{$\bf\diamond$} Pang et al.~\cite{pang2022robustness}$^{\ddagger}$ & 63.66 & 31.08 \\
    \checkmark & \xmark & \textcolor{green}{$\bf\diamond$} Zhou et al.~\cite{zhounasty} & 64.02 & 28.82 & \textcolor{green}{$\bf\diamond$} Jin et al.~\cite{jin2025s2o}$^{\ddagger}$  & 63.77 & 31.29 \\

    \cdashline{1-8}
    
    \checkmark & \xmark & \textcolor{blue}{$\bf\diamond$} Shi et al.~\cite{shi2021online} & 52.91 & 32.38 & \textcolor{blue}{$\bf\diamond$} Shi et al.~\cite{shi2021online} & 61.01 & 34.37 \\
    \xmark & \checkmark & \textcolor{blue}{$\bf\diamond$} Mao et al.~\cite{mao2021adversarial} & -- & 25.45 & \textcolor{blue}{$\bf\diamond$} Mao et al.~\cite{mao2021adversarial} & -- & 33.16 \\
    \xmark & \checkmark & \textcolor{blue}{$\bf\diamond$} Kulkarni et al.~\cite{kulkarni2024interpretability} & 65.37 & 32.55 & \textcolor{blue}{$\bf\diamond$} Hwang et al.~\cite{hwang2023aid}* & 64.73 & 32.86 \\
    
    \color{lightgray}\xmark & \color{lightgray}\xmark & \color{lightgray}\hphantom{\textcolor{blue}{$\bf\diamond$}} Gaussian Blur & \color{lightgray}49.46 & \color{lightgray}20.65 & \color{lightgray}\hphantom{\textcolor{blue}{$\bf\diamond$}} Gaussian Blur & \color{lightgray}50.41 & \color{lightgray}20.56 \\
    
    \xmark & \xmark & \textcolor{blue}{$\bf\diamond$} \textbf{GS} & 30.42 & 34.35 & \textcolor{blue}{$\bf\diamond$} \textbf{GS} & 26.56 & 32.06 \\
    \cellcolor{my_color}\xmark & \cellcolor{my_color}\xmark & \cellcolor{my_color}\textcolor{blue}{$\bf\diamond$} \textbf{GS+AP (ZeroPur)} & \cellcolor{my_color}52.45 & \cellcolor{my_color}\textbf{41.42} & \cellcolor{my_color}\textcolor{blue}{$\bf\diamond$} \textbf{GS+AP (ZeroPur)} & \cellcolor{my_color}\textbf{64.88} & \cellcolor{my_color}\textbf{34.58} \\
    \bottomrule
\end{tabular}
\label{tab:cifar-10}
\end{table*}

\begin{table}[!t]
\centering
\caption{Performance comparison with external model-based purification (EBP)  methods on CIFAR-10 under AutoAttack $\ell_{\infty} (\epsilon= 8 /255)$.  \textcolor{red}{$\bf\diamond$} and \textcolor{blue}{$\bf\diamond$} denote EBP and ABP methods, respectively. $^{\dagger}$ indicates that ZeroPur is applied to a victim classifier trained with strong data augmentation.}
\label{tab:ebp-cifar10-inf}
\fontsize{8.5}{9}\selectfont 
\renewcommand{\tabcolsep}{5pt}
\renewcommand{\arraystretch}{1.0}
\begin{tabular}{cclcc}
\toprule
\multicolumn{2}{c}{Require Training} &  \multicolumn{3}{c}{WideResNet-28-10} \\
\cmidrule(lr){1-2} \cmidrule(lr){3-5} 
Cls & Ext &  Method & SA(\%) & RA(\%) \\
\midrule
\xmark & \checkmark & \textcolor{red}{$\bf\diamond$} Yoon et al.~\cite{yoon2021adversarial} & 85.66 & 59.53 \\
\xmark & \checkmark & \textcolor{red}{$\bf\diamond$} Ughini et al.~\cite{ughini2022trust} & -- & 59.57 \\
\xmark & \checkmark & \textcolor{red}{$\bf\diamond$} Nie et al.~\cite{nie2022diffusion} & 89.02 & 70.64 \\
\xmark & \checkmark & \textcolor{red}{$\bf\diamond$} Lee et al.~\cite{lee2023robust} & 90.16 & 70.47 \\
\xmark & \checkmark & \textcolor{red}{$\bf\diamond$} Lin et al.~\cite{lin2024adversarial} & 90.62 & 72.85 \\
\xmark & \checkmark & \textcolor{red}{$\bf\diamond$} Bai et al.~\cite{bai2024diffusion} & 91.41 & 77.08 \\
\cdashline{1-5}
\rowcolor{my_color}\xmark & \xmark & \textcolor{blue}{$\bf\diamond$} \textbf{ZeroPur} & \textbf{91.81} & 68.60 \\
\rowcolor{my_color}\xmark & \xmark & \textcolor{blue}{$\bf\diamond$} \textbf{ZeroPur$^{\dagger}$} & 85.02 & \textbf{82.76} \\
\bottomrule
\end{tabular}

\end{table}

\begin{table}[!t]
\centering
\caption{Performance comparison with adversarial training (AT), auxiliary-based purification (ABP), and external model-based purification (EBP) methods on ImageNet-1K under AutoAttack $\ell_{\infty} (\epsilon= 4 /255)$. \textcolor{green}{$\bf\diamond$}, \textcolor{blue}{$\bf\diamond$}, and \textcolor{red}{$\bf\diamond$} denote AT, ABP, and EBP methods, respectively.}
\label{fig:imagenet}
\fontsize{8.5}{9}\selectfont 
\renewcommand{\tabcolsep}{5pt}
\renewcommand{\arraystretch}{1.0}
\begin{tabular}{ccclcc}
\toprule
\multicolumn{3}{c}{Require Training} & \multicolumn{3}{c}{ResNet-50} \\
\cmidrule(lr){1-3}\cmidrule(lr){4-6} 
Cls & Aux & Ext & Method & SA(\%) & RA(\%) \\
\midrule
\checkmark & \xmark & \xmark & \textcolor{green}{$\bf\diamond$} Wong et al.~\cite{wong2020fast} & 55.62 & 26.24 \\
\checkmark & \xmark & \xmark & \textcolor{green}{$\bf\diamond$} Salman et al.~\cite{salman2020adversarially} & 64.02 & 34.96 \\
\checkmark & \xmark & \xmark & \textcolor{green}{$\bf\diamond$} Bai et al.~\cite{bai2021transformers} & 67.38 & 35.51 \\
\checkmark & \xmark & \xmark & \textcolor{green}{$\bf\diamond$} Chen et al.~\cite{chen2024data} & 68.76 & 40.60 \\
\cdashline{1-6}
\xmark & \checkmark & \xmark & \textcolor{blue}{$\bf\diamond$} Mao et al.~\cite{mao2021adversarial} & -- & 31.32 \\
\xmark & \checkmark & \xmark & \textcolor{blue}{$\bf\diamond$} Kulkarni et al.~\cite{kulkarni2024interpretability} & 64.10 & 37.36 \\
\cdashline{1-6}
\xmark & \xmark & \checkmark & \textcolor{red}{$\bf\diamond$} Nie et al.~\cite{nie2022diffusion} & 67.79 & 40.93 \\
\xmark & \xmark & \checkmark & \textcolor{red}{$\bf\diamond$} Lee et al.~\cite{lee2023robust} & 70.74 & 46.31 \\
\xmark & \xmark & \checkmark & \textcolor{red}{$\bf\diamond$} Song et al.~\cite{song2024mimicdiffusion} & 66.92 & 62.16 \\
\xmark & \xmark & \checkmark & \textcolor{red}{$\bf\diamond$} Pei et al.~\cite{pei2025diffusion} & \textbf{71.88} & 59.77 \\

\cdashline{1-6}
{\color{lightgray}\xmark} & {\color{lightgray}\xmark} & {\color{lightgray}\xmark} & {\color{lightgray}\hphantom{\textcolor{blue}{$\bf\diamond$}} Gaussian Blur} & {\color{lightgray}71.29} & {\color{lightgray}60.35} \\
\xmark & \xmark & \xmark & \textcolor{blue}{$\bf\diamond$} \textbf{GS} & 63.38 & 65.72 \\
\rowcolor{my_color}\xmark & \xmark & \xmark & \textcolor{blue}{$\bf\diamond$} \textbf{GS+AP (ZeroPur)} & \textbf{71.88} & \textbf{67.09} \\  
\bottomrule
\end{tabular}
\end{table}

\subsection{Comparison with Existing Methods}
For adversarial defense, adversarial training (AT) involves adversarial examples in the classifier training, and auxiliary-based purification (ABP) introduces an auxiliary function to cooperate with the victim classifier. These methods do not rely on external generative models but still necessitate training auxiliary functions or retraining classifiers. External model-based purification (EBP) fine-tunes or retrains an external generative model to remove adversarial perturbations in adversarial images. Since the performance of EBP is ensured by the modeling capability of generative models, it typically outperforms ABP. In this section, for a fair comparison, we will first compare our method with both AT and ABP and then compare it with EBP. In Table~\ref{tab:cifar-10}, ~\ref{tab:ebp-cifar10-inf}, and ~\ref{fig:imagenet}, the terms ‘Ext', ‘Aux', and ‘Cls' under ‘Require Training' indicate whether the methods require retraining external generative models, auxiliary functions, or victim classifiers.

\textbf{Comparison with AT $\&$ ABP.} Table~\ref{tab:cifar-10} reports the standard and robust accuracy against AutoAttack $\ell_{\infty}$ ($\epsilon = 8/255$). ZeroPur achieves consistent and significant improvements in robust accuracy, particularly on ResNet-18, with robust accuracies of $69.62\%$ and $41.42\%$ on CIFAR-10 and CIFAR-100, respectively. Notably, several strong AT baselines in Table~\ref{tab:cifar-10} leverage large-scale synthetic data during training to further enhance robustness (marked by $\ddagger$), whereas ZeroPur still achieves outstanding robustness without requiring any additional data or classifier retraining. Additionally, we report the comparison on ImageNet-1K in Table~\ref{fig:imagenet}, where ZeroPur improves robust accuracy by $26.49\%$ compared to AT and ABP methods, representing a substantial gain. These results demonstrate that ZeroPur effectively guides adversarial images toward the natural image manifold. Furthermore, our method performs slightly better on smaller architectures (e.g., ResNet-18) than on larger ones (e.g., WideResNet-28-10). This counter-intuitive result occurs because the complex parameters of larger models act as stronger constraints during the purification process, which inadvertently increases the difficulty of effectively mapping adversarial images back to the natural image manifold.

\textbf{Comparison with EBP.} Although generative model-based purification methods achieve outstanding performance due to their large scale and computational cost, our method still achieves comparable or state-of-the-art performance. Table~\ref{tab:ebp-cifar10-inf} and Table~\ref{fig:imagenet} report comparisons between ZeroPur and representative EBP methods. On CIFAR-10, ZeroPur achieves robust accuracy comparable to existing EBP defenses. For completeness, we additionally report ZeroPur$^{\dagger}$ (see Section~\ref{ab}), where the same training-free purification pipeline is applied to a victim classifier trained with stronger data augmentation. Under this evaluation setting, ZeroPur$^{\dagger}$ further improves the performance among EBP methods.
Notably, on the highly challenging ImageNet-1K dataset, ZeroPur also achieves state-of-the-art robust accuracy. Specifically, our full pipeline, GS+AP (ZeroPur), obtains a robust accuracy of $67.09\%$ and a standard accuracy of $71.88\%$. This represents significant improvements of $4.93\%$ and $4.96\%$, respectively, compared to Mimicdiffusion~\cite{song2024mimicdiffusion}, the leading EBP method in terms of robustness. This result highlights the efficacy of our method in achieving superior performance without relying on large, resource-intensive external models.

To improve comparison transparency, Gaussian Blur is explicitly included as a preprocessing baseline. As shown in Table~\ref{tab:cifar-10} and ~\ref{fig:imagenet}, Gaussian Blur consistently underperforms ZeroPur, indicating that the gains of ZeroPur cannot be attributed solely to preprocessing but arise from feature-guided manifold projection. Notably, although Gaussian Blur achieves competitive performance on ImageNet-1K and even surpasses several existing defenses, this does not imply that blurring itself performs purification. Instead, blur only provides a coarse manifold-oriented initialization, which ZeroPur further refines through iterative optimization for effective purification.

\begin{figure}[!t]
    \centering
    \subfloat[ResNet-18]{%
        \includegraphics[width=0.48\linewidth]{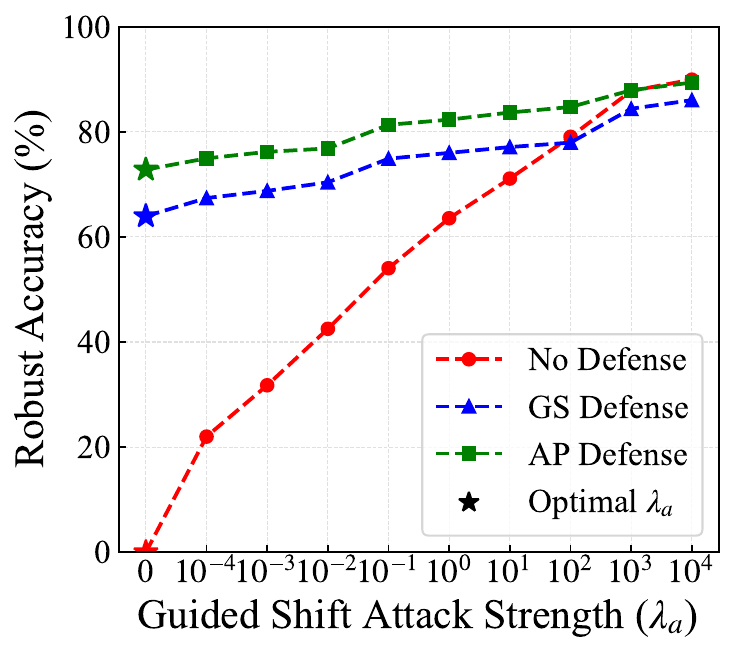}
    }
    \hfill
    \subfloat[WideResNet-28-10]{%
        \includegraphics[width=0.48\linewidth]{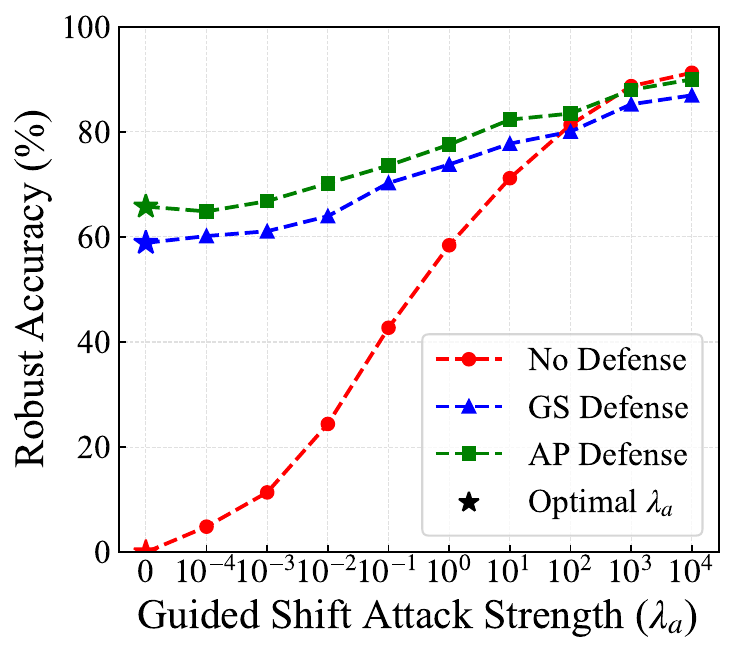}
    }
\caption{Robust accuracy of defense-aware attacks under different $\lambda_{a}$ values.
We vary $\lambda_{a}$ from 0 to $10^4$ and report results on two standard models (ResNet-18 and WideResNet-28-10) on CIFAR-10. Increasing $\lambda_{a}$ not only weakens the gain of our method over the baseline (reducing the difference) but also decreases the attack effectiveness against standard models. The optimal attack occurs at $\lambda_{a}$ = 0, i.e., the standard attack without targeting the GS defense.}
\label{fig05}
\end{figure}

\subsection{Result on Adaptive Attack}
\label{defense aware}
To more realistically demonstrate the robustness of our method, we consider the scenario where the adversary has full knowledge of our defense strategy. Intuitively, in order to undermine the effectiveness of our method, the adversary needs to achieve two objectives simultaneously:
(1) generate adversarial examples that can cause the model to make incorrect predictions;
(2) ensure that, after applying a blurring process, these adversarial examples no longer provide a purification benefit.
Mathematically, this can be formulated as:
\begin{equation}
\begin{cases}
\max \ \ell(f \circ g(\bm{x}_{\mathrm{adv}}), y) \\
\min\ -d(f(\bm{x}_{\mathrm{adv}}), f(\bm{x}_{\mathrm{adv}}^{\prime}))
\end{cases}
\, \text{s.t.} \ \ \Vert \bm{x}_{\mathrm{adv}}-\bm{x} \Vert \leq \epsilon
\end{equation}
here, $d(\cdot, \cdot)$ denotes the cosine similarity, and $\bm{x}_{\mathrm{adv}}^{\prime}$ is the blurred counterpart of $\bm{x}_{\mathrm{adv}}$. Following CW~\cite{carlini2017towards}, in our implementation, we optimize   
$\bm{x}_{\mathrm{adv}}$ by maximizing the following formula:
\begin{equation}
\mathcal{L}_{a} = \ell(f \circ g(\bm{x}_{\mathrm{adv}}), y) + \lambda_{a}d(f(\bm{x}_{\mathrm{adv}}), f(\bm{x}_{\mathrm{adv}}^{\prime}))
\end{equation}
where $\lambda_{a}$ controls the degree to which the adversary attacks our GS. The above equation shows that the adversary attempts to generate adversarial examples by bypassing our guided shift; if the generated examples have the same embedding as the blurred ones, our GS will no longer provide any guidance. However, the adversary must balance such a multi-objective attack, which results in adversarial examples that are not truly adversarial when the attack is focused on our method.

Fig.~\ref{fig05} shows the performance of the adaptive attack on CIFAR-10. We use a PGD attack with 40 steps and a perturbation bound of $L_\infty = 8/255$, varying $\lambda_a$ from 0 to $10^4$, where $\lambda_a = 0$ corresponds to the standard PGD attack without considering our defense strategy. The results indicate that as $\lambda_a$ increases, more attack budget is allocated to targeting the GS defense, which reduces the advantage of our method over the baseline (smaller difference). However, at the same time, the attack effectiveness against the baseline also drops significantly (curve rises). Although adaptive attacks can diminish the extra gain brought by our method, doing so greatly sacrifices the original classification attack success rate. Notably, unlike prior work~\cite{mao2021adversarial}, our baseline is a non-robust standard model, which means a much larger $\lambda_a$ is required to completely neutralize our defense relative to the baseline. This, however, entirely contradicts the goal of adversarial attacks, making it unworthy for the adversary to account for our defense. Therefore, even under defense aware attack, our method consistently demonstrates effective protection compared to the baseline.

\begin{table}[!t]
\centering
\caption{RA ($\%$) against BPDA $\ell_{\infty}$ ($\epsilon=8/255$) on CIFAR-10.}
\label{bpda}
\fontsize{8.5}{9}\selectfont 
\renewcommand{\tabcolsep}{5pt}
\renewcommand{\arraystretch}{1.0}
\begin{tabular}{ccccc}
\toprule
Setting & Model & PGD-10 & PGD-20 & PGD-40 \\
\midrule
\multirow{2}{*}{Vanilla} 
& WRN & \textbf{50.61}     &  \textbf{50.71}     & \textbf{50.54} \\
& R18 & 34.44   &   34.88   & 32.18 \\
\midrule
\multirow{2}{*}{Base}
& WRN & 34.16 & 33.38 & 34.42 \\
& R18 & \textbf{39.88} & \textbf{39.90} & \textbf{39.48} \\
\midrule
\multirow{2}{*}{Strong}
& WRN & 68.84 & 69.45 & 68.50 \\
& R18 & \textbf{70.65} & \textbf{70.45} & \textbf{70.43} \\
\bottomrule
\end{tabular}
\end{table}

\subsection{Result on other Attack}
\subsubsection{ZeroPur defense against attacks that bypass the purification module} The efficiency of purification methods can be challenged by the BPDA attack that approximates gradients, bypassing the purification module. Table~\ref{bpda} reports the performance of ZeroPur defense against BPDA attack. We can observe that the robust accuracy of ZeroPur decreases, but remains stable, indicating that ZeroPur provides acceptable defense.

\begin{figure}[!t]
    \centering
    \subfloat[Blur vs. TVM\label{reg:blur}]{%
        \includegraphics[width=0.48\linewidth]{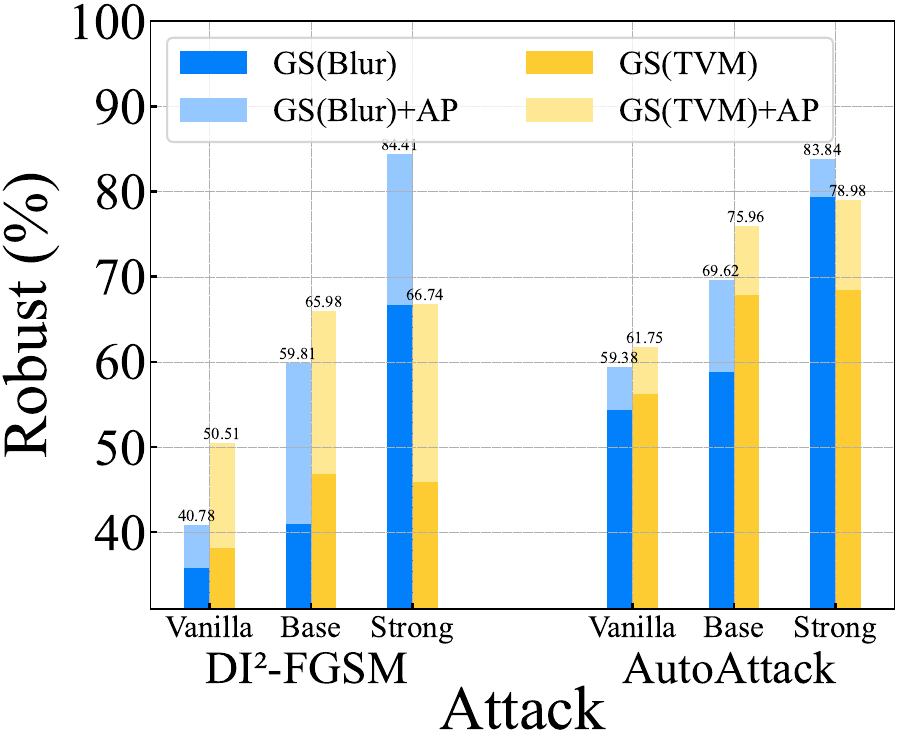}
    }
    \hfill
    \subfloat[Ablation of PR\label{reg:pr}]{%
        \includegraphics[width=0.48\linewidth]{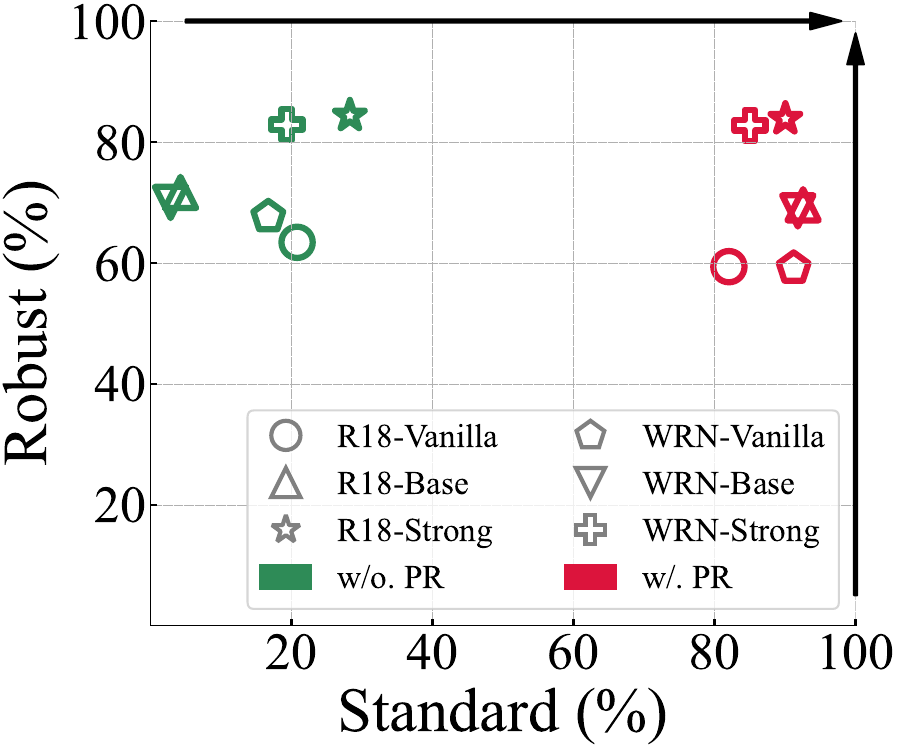}
    }
\caption{(a) Robust accuracy against DI$^{2}$-FGSM and AutoAttack on CIFAR-10 by Blur and TVM. (b) Ablation study of perceptual regularization on CIFAR-10.}
\label{reg}
\end{figure}

\subsubsection{ZeroPur defense against adversarial attacks robust to blurring} ZeroPur employs blurring to project adversarial images back onto the natural image manifold. Although certain attacks, such as DI$^{2}$-FGSM\cite{xie2019improving}, exhibit inherent resistance to blurring, they are still not entirely immune to ZeroPur. As shown in Fig.~\ref{reg}\textcolor{blue}{(a)}, replacing the blurring operator with TVM~\cite{guo2017countering} leads to improved purification. The robust accuracy against AutoAttack increases to $75.96\%$ without training, which outperforms the optimal performance of GS (Blur) +AP by $6.34\%$. This result also demonstrates that ZeroPur is not limited to blurring and can generalize to other operators.

\subsection{Ablation Study}
To validate the effectiveness of perceptual regularization (PR) in \equref{AP}, we conducted ablation studies, and the experimental results are illustrated in Fig.~\ref{reg}\textcolor{blue}{(b)}. The results compare models trained with and without PR on different architectures, including ResNet-18 (R18) and WideResNet-28-10 (WRN). As shown in the figure, models without PR (green markers) generally achieve lower standard accuracy, while incorporating PR (red markers) significantly improves standard accuracy without sacrificing robust accuracy. This phenomenon is consistent across different training settings, such as Vanilla, Base, and Strong, confirming that PR can effectively enhance model performance in the adaptive projection stage.

\begin{table*}[t]
\centering
\caption{SA / RA (\%) on CIFAR-10 with different augmentations under varying blurring operators. The \textbf{bold} indicates the best harmonic mean of SA and RA, measured by the F1 Score ($F1 = 2 \cdot \frac{SA \cdot RA}{SA + RA}$).}
\fontsize{8.5}{9}\selectfont 
\renewcommand{\tabcolsep}{2pt}
\renewcommand{\arraystretch}{1.1}
\begin{tabular}{>{\centering\arraybackslash}p{1.3cm} >{\centering\arraybackslash}p{1.7cm} *{3}{>{\centering\arraybackslash}p{2.2cm}} c *{3}{>{\centering\arraybackslash}p{2.2cm}}}
\toprule
\multirow{2}{*}{Operator} & \multirow{2}{*}{Level} & \multicolumn{3}{c}{ResNet-18} & & \multicolumn{3}{c}{WRN-28-10} \\
\cmidrule(lr){3-5} \cmidrule(lr){7-9}
& & Vanilla & Base & Strong & & Vanilla & Base & Strong \\
\midrule

\multirow{3}{*}{Median} 
& $3 \times 3$     
& \fOneCellRange{82.04}{59.38}{0.3287}{0.6889} & \fOneCellRange{92.34}{72.27}{0.6826}{0.8108} & \fOneCellRange{90.90}{60.74}{0.5020}{0.8683} &
& \fOneCellRange{91.22}{59.16}{0.3521}{0.7177} & \fOneCellRange{91.81}{67.56}{0.4494}{0.7866} & \fOneCellRange{90.65}{70.01}{0.5549}{0.8387} \\

& $5 \times 5$     
& \fOneCellRange{83.73}{34.17}{0.3287}{0.6889} & \fOneCellRange{91.87}{64.81}{0.6826}{0.8108} & \fOneCellRange{90.92}{44.63}{0.5020}{0.8683} &
& \fOneCellRange{91.06}{39.81}{0.3521}{0.7177} & \fOneCellRange{91.34}{58.41}{0.4494}{0.7866} & \fOneCellRange{88.97}{50.49}{0.5549}{0.8387} \\

& $7 \times 7$     
& \fOneCellRange{83.66}{27.25}{0.3287}{0.6889} & \fOneCellRange{91.56}{54.41}{0.6826}{0.8108} & \fOneCellRange{90.87}{34.68}{0.5020}{0.8683} &
& \fOneCellRange{90.97}{28.94}{0.3521}{0.7177} & \fOneCellRange{90.34}{48.26}{0.4494}{0.7866} & \fOneCellRange{87.70}{40.58}{0.5549}{0.8387} \\

\hline

\multirow{3}{*}{Gaussian} 
& $\sigma = 0.6$  
& \fOneCellRange{83.57}{38.82}{0.3287}{0.6889} & \fOneCellRange{91.64}{65.43}{0.6826}{0.8108} & \fOneCellRange{90.72}{64.92}{0.5020}{0.8683} &
& \fOneCellRange{91.09}{53.21}{0.3521}{0.7177} & \fOneCellRange{90.02}{58.31}{0.4494}{0.7866} & \fOneCellRange{90.68}{73.79}{0.5549}{0.8387} \\

& $\sigma = 1.2$ 
& \fOneCellRange{83.63}{26.20}{0.3287}{0.6889} & \fOneCellRange{92.56}{69.62}{0.6826}{0.8108} & \fOneCellRange{90.05}{83.84}{0.5020}{0.8683} &
& \fOneCellRange{91.14}{27.97}{0.3521}{0.7177} & \fOneCellRange{91.81}{68.80}{0.4494}{0.7866} & \fOneCellRange{85.02}{82.76}{0.5549}{0.8387} \\

& $\sigma = 1.8$  
& \fOneCellRange{83.69}{20.45}{0.3287}{0.6889} & \fOneCellRange{92.36}{60.67}{0.6826}{0.8108} & \fOneCellRange{90.79}{67.89}{0.5020}{0.8683} &
& \fOneCellRange{51.26}{26.82}{0.3521}{0.7177} & \fOneCellRange{35.95}{59.91}{0.4494}{0.7866} & \fOneCellRange{80.27}{78.82}{0.5549}{0.8387} \\

\bottomrule
\end{tabular}
\label{F1}
\end{table*}

\begin{figure}[!t]
    \centering
    \subfloat[ResNet-18]{%
        \includegraphics[width=0.48\linewidth]{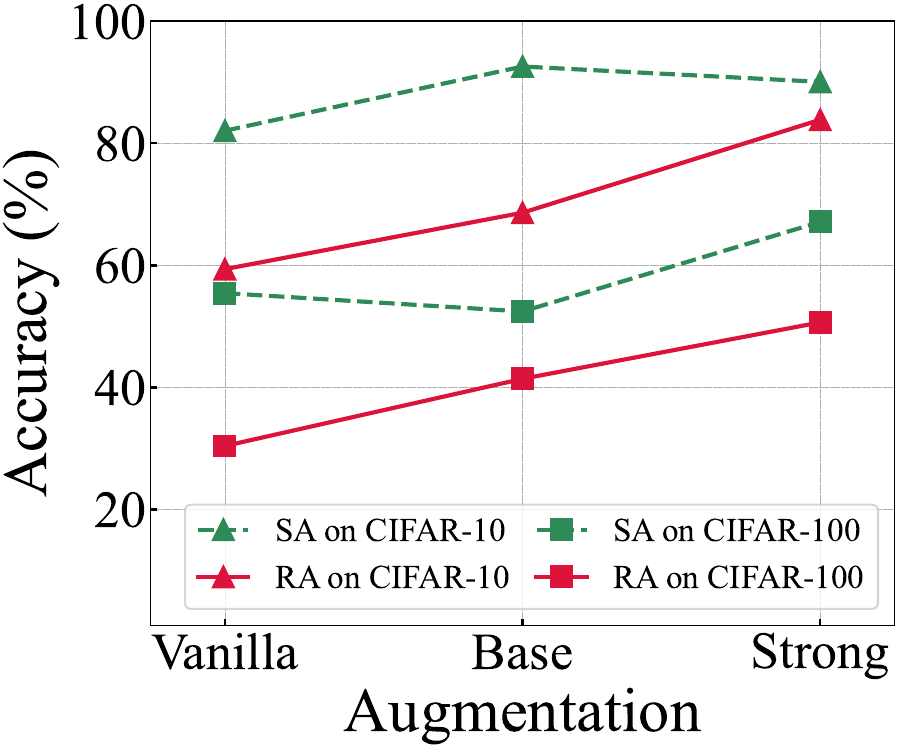}
    }
    \hfill
    \subfloat[WideResNet-28-10]{%
        \includegraphics[width=0.48\linewidth]{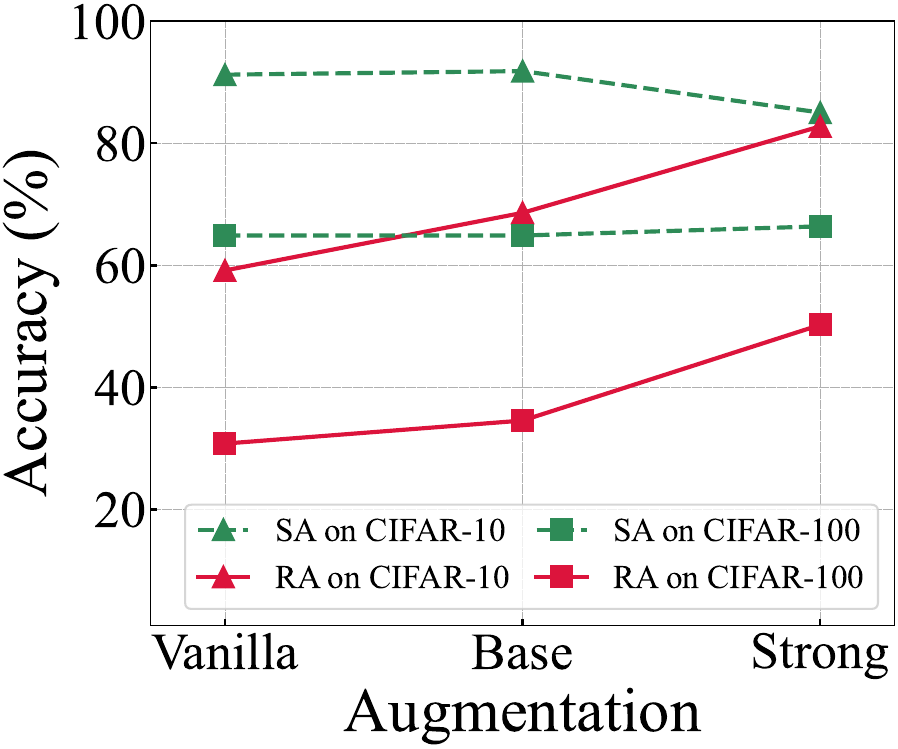}
    }
\caption{SA and RA (\%) of classifiers trained with three-level data augmentation on (a) ResNet-18 and (b) WideResNet-28-10.}
\label{fig06}
\end{figure}


\subsection{Sensitivity Analysis}
\label{ab}
In our method, the strength of the blur operator naturally affects purification performance. Therefore, we conduct a sensitivity analysis to investigate the interaction between blur strength and the training recipe of the victim classifier. Specifically, we evaluate ZeroPur on pretrained victim classifiers trained with three augmentation strategies (‘Vanilla’, ‘Base’, and ‘Strong’), using Median filters ($3\times3$, $5\times5$, $7\times7$) and Gaussian blurs ($\sigma=0.6$, $1.2$, $1.8$). It is important to note that these augmentation strategies are used only to obtain alternative victim classifiers for analysis purposes. The ZeroPur purification pipeline itself remains unchanged and requires no additional training.

Table~\ref{F1} reports the resulting standard and robust accuracies. When the victim classifier is trained without strong augmentation, aggressive blur operations tend to produce low-quality embeddings, reducing the effectiveness of Guided Shift. In contrast, classifiers trained with strong augmentation are inherently more tolerant to blurred inputs, allowing ZeroPur to benefit from stronger blur strength and achieve higher robust accuracy. For commonly used pretrained classifiers (e.g., standard PyTorch models), moderate blur strengths generally provide the best trade-off between standard and robust accuracy, which is quantitatively measured by their harmonic mean (i.e., the F1 score). Accordingly, ZeroPur adopts a Median filter ($3\times3$) for ‘Vanilla’ classifiers and Gaussian blur ($\sigma=1.2$) for both ‘Base’ and ‘Strong’ classifiers throughout the paper.

Fig.~\ref{fig06} further illustrates the standard and robust accuracies on CIFAR-10 and CIFAR-100 using ResNet-18 and WideResNet-28-10. Consistent with the observations in Table~\ref{F1}, stronger data augmentation during classifier training improves robustness to blurred inputs, which in turn enhances the effectiveness of ZeroPur.

\begin{table}[!t]
\centering
\caption{Generalization performance of ZeroPur on CIFAR-10 (ResNet-18) by varying the perturbation radius $\epsilon_{pfy}$ and the attack radius $\epsilon_{atk}$. The values represent SA / RA (\%).}
\label{tab:zeropur_epsilon_generalization}

\fontsize{8.5}{9}\selectfont 
\renewcommand{\tabcolsep}{5pt}
\renewcommand{\arraystretch}{1.0}
\begin{tabular}{cccc}
\toprule
\multirow{2}{*}{$\epsilon_{pfy}$} & \multicolumn{3}{c}{$\epsilon_{atk}$} \\
\cmidrule(lr){2-4}
& $4 / 255$ & $8 / 255$ & $16 / 255$ \\
\cmidrule(lr){1-4} 
$0.50 \epsilon_{atk}$  & 92.05/ 30.15 & 92.71/ 34.71 & 92.55/ 24.60 \\
$0.75 \epsilon_{atk}$ & 89.95/ 70.20 & 92.48/ 65.87 & 92.56/ 45.18 \\
$1.00 \epsilon_{atk}$    & 88.64/ 74.88 & 92.54/ 70.27 & 92.28/ 53.11 \\
$1.25 \epsilon_{atk}$ & 87.57/ 73.48 & 92.56/ 69.62 & 92.27/ 54.77 \\
$1.50 \epsilon_{atk}$  & 87.61/ 70.71 & 92.43/ 67.60 & 91.86/ 54.72 \\
$2.00 \epsilon_{atk}$    & 87.31/ 66.56 & 92.35/ 65.96 & 91.39/ 54.35 \\
\bottomrule
\end{tabular}
\end{table}

\subsection{Generalization Across Attack and Purification Radii}
To validate the generalization of ZeroPur, we analyze model performance by varying the pixel perturbation radius $\epsilon_{pfy}$ relative to the attack radius $\epsilon_{atk}$, with results reported in Table~\ref{tab:zeropur_epsilon_generalization}.

The experimental results strongly support the condition $\epsilon_{pfy} \ge \epsilon_{atk}$ as essential for optimal robust performance. Setting the defense radius $\epsilon_{pfy}$ to be greater than or equal to the attack radius $\epsilon_{atk}$ ensures that the defense mechanism operates within a sufficient perturbation space, which is critical for accurately estimating and effectively eliminating the adversarial direction. The robust accuracy remains high and stable across different $\epsilon_{pfy}$ values once this condition is met (e.g., at $\epsilon_{atk}=8/255$, the robust accuracy is consistently high from $\epsilon_{pfy}=1 \epsilon_{atk}$ to $2 \epsilon_{atk}$).

Conversely, when $\epsilon_{pfy}$ is smaller than $\epsilon_{atk}$, the defense space is limited, making it insufficient for effective perturbation elimination. Given that real-world adversarial attacks are typically constrained by a small $\epsilon_{atk}$ for stealth, and the defense radius $\epsilon_{pfy}$ is user-defined and can be set generously, satisfying the necessary condition $\epsilon_{pfy} \ge \epsilon_{atk}$ is highly practical and advantageous. Therefore, we recommend setting a sufficiently large $\epsilon_{pfy}$ to ensure stable and high robust accuracy. Based on this analysis, we set $\epsilon_{pfy}$ to $1.25 \times \epsilon_{atk}$ as the optimal baseline configuration.

\begin{figure*}[!t]
\centering
\includegraphics[width=\linewidth]{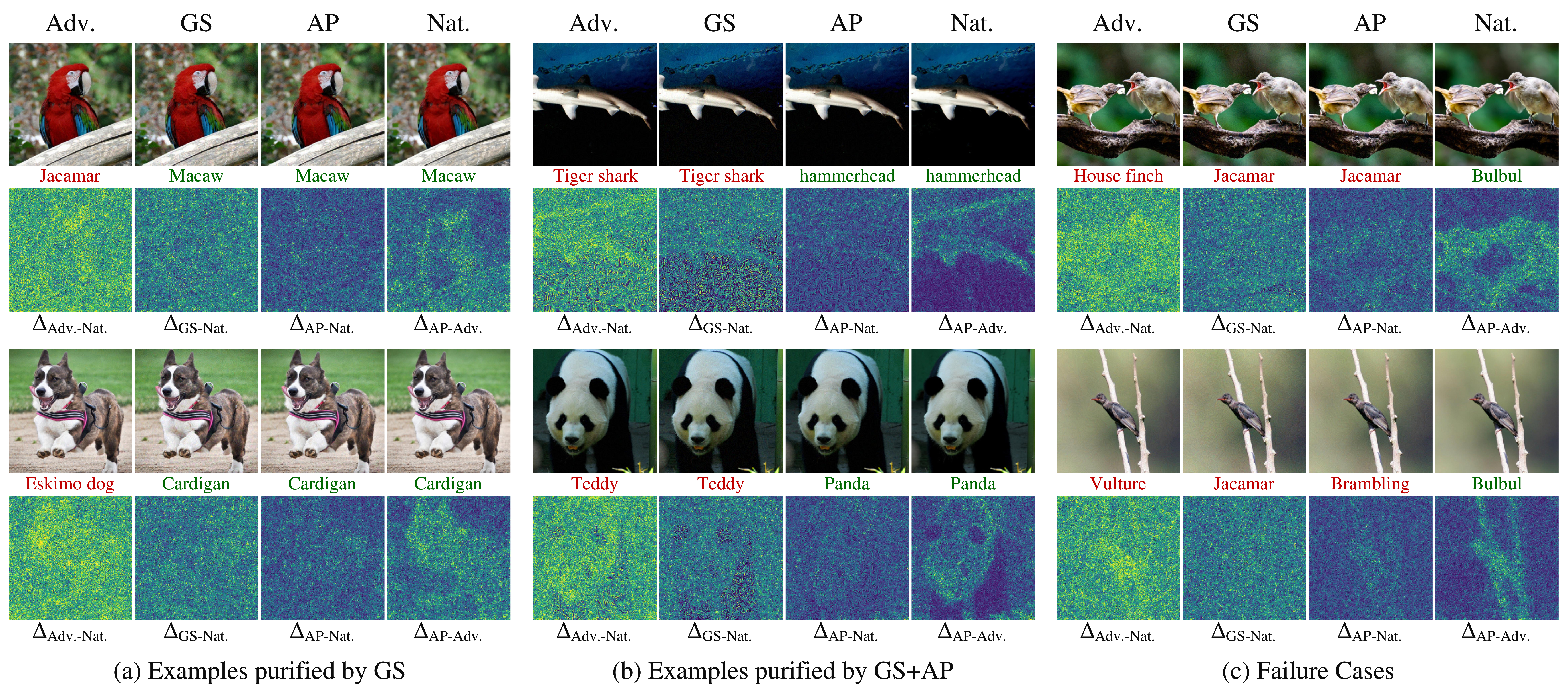}
\caption{Visual examples of ZeroPur against $\ell_{\infty}$ threat model ($\epsilon=4/255$) on ImageNet-1K. The \textcolor{darkred}{\textbf{red}} denotes error prediction and the \textcolor{darkgreen}{\textbf{green}} denotes correct prediction. $\Delta_{\mathrm{A}\text{-}\mathrm{B}}$ represents the normalized absolute difference between A and B.}
\label{fig:visualization}
\end{figure*}

\subsection{Inference Efficiency}
To evaluate the practical viability of ZeroPur, we report the inference throughput in Table \ref{tab:inference_cost}. Experiments were conducted on a single NVIDIA RTX A6000 GPU, with statistics for CIFAR-10 and CIFAR-100 merged due to their nearly identical overheads. Beyond its zero-setup and domain-agnostic nature, ZeroPur exhibits superior efficiency; unlike Adversarial Training (AT) or ABP-based methods that necessitate exhaustive training phases, ZeroPur’s efficiency advantage is intrinsic and immediate. Notably, ZeroPur reduces inference time by over two orders of magnitude compared to diffusion-based methods like Mimicdiffusion~\cite{song2024mimicdiffusion} (which requires 156s per ImageNet image on a superior RTX 4090, as reported in its original paper). This performance gap, maintained despite the hardware disadvantage of the RTX A6000, strongly underscores the low-latency superiority of ZeroPur.

\begin{table}[!t]
  \centering
  \caption{Inference throughput (Images Per Second, IPS) of ZeroPur across different datasets and models. Tested on a single NVIDIA RTX A6000 GPU. Data for CIFAR-10 and CIFAR-100 are merged due to nearly identical inference times.}
  \label{tab:inference_cost}
  \fontsize{8.5}{9}\selectfont 
\renewcommand{\tabcolsep}{5pt}
\renewcommand{\arraystretch}{1.0}
  \begin{tabular}{lccc}
    \toprule
    \multirow{2}{*}{Phase} & \multicolumn{2}{c}{CIFAR-10 \& 100} & ImageNet-1K \\
    \cmidrule(lr){2-3} \cmidrule(lr){4-4}
    & ResNet-18 & WRN-28-10 & ResNet-50 \\
    \midrule
    Guided Shift (GS)        & 303.85 & 71.11 & 14.95 \\
    Adaptive Projection (AP) & 33.07  & 4.21  & 2.36  \\
    GS+AP (ZeroPur)          & 29.81  & 3.97  & 2.04  \\
    \bottomrule
  \end{tabular}
\end{table}

\subsection{Visualization}
To gain a deeper understanding of our method, we present intuitive visualizations and differences resulting from applying the two-step ZeroPur process to adversarial examples. As shown in Fig.~\ref{fig:visualization}, the progression from adversarial images to GS-purified images and then to AP-purified images shows decreasing differences from natural images (gradually transitioning from yellow to blue), indicating the effective removal of adversarial noise. We observe that in cases where GS successfully purifies adversarial samples, the difference between GS-purified and natural images (denoted as $\Delta_{\text{GS-Nat.}}$) displays scattered, non-semantic noise. In contrast, for cases where GS fails to purify,  the difference $\Delta_{\text{GS-Nat.}}$ still shows semantic perturbations. Furthermore, the similarity between $\Delta_{\text{AP-Adv.}}$ and $\Delta_{\text{Adv.-Nat.}}$ suggests that our method (GS+AP, i.e., ZeroPur) focuses purification mainly on regions influenced by adversarial perturbations. In summary, the reduced difference between purified and natural images, along with the increased difference from adversarial images, demonstrates the effectiveness of our method. However, some failure cases still exist. It is not difficult to observe that the predictions of adversarial samples and GS are already inconsistent at the initial stage. This highlights the importance of the initial purification direction, which may be key to ZeroPur's success.

\section{Conclusion}
\label{sec:conclusion}
We proposed ZeroPur, a succinct and training-free adversarial purification method grounded in the natural image manifold hypothesis. ZeroPur integrates Guided Shift (GS) and Adaptive Projection (AP), which collaboratively guide adversarial examples back toward the natural manifold without retraining external models, auxiliary functions, or victim classifiers. Extensive experiments on CIFAR-10, CIFAR-100, and ImageNet-1K demonstrate that ZeroPur not only surpasses prior state-of-the-art adversarial training and auxiliary-based purification methods but also achieves performance comparable to external model-based approaches. Although ZeroPur is simple and efficient, its blur-based strategy remains less effective against strong adaptive attacks, such as BPDA evaluated in this paper, likely due to the absence of external models that limits its defensive capacity. To overcome this challenge, it is necessary to design modules that emulate recent advances in diffusion-based purification, thereby enabling more effective defense against BPDA and similar attacks. Overall, ZeroPur underscores the promise of lightweight, training-free defenses for practical adversarial robustness.

\bibliographystyle{IEEEtran}
\bibliography{egbib}

\vspace{-1em}
\begin{IEEEbiography}[{\includegraphics[width=0.91in,height=1.25in,clip,keepaspectratio]{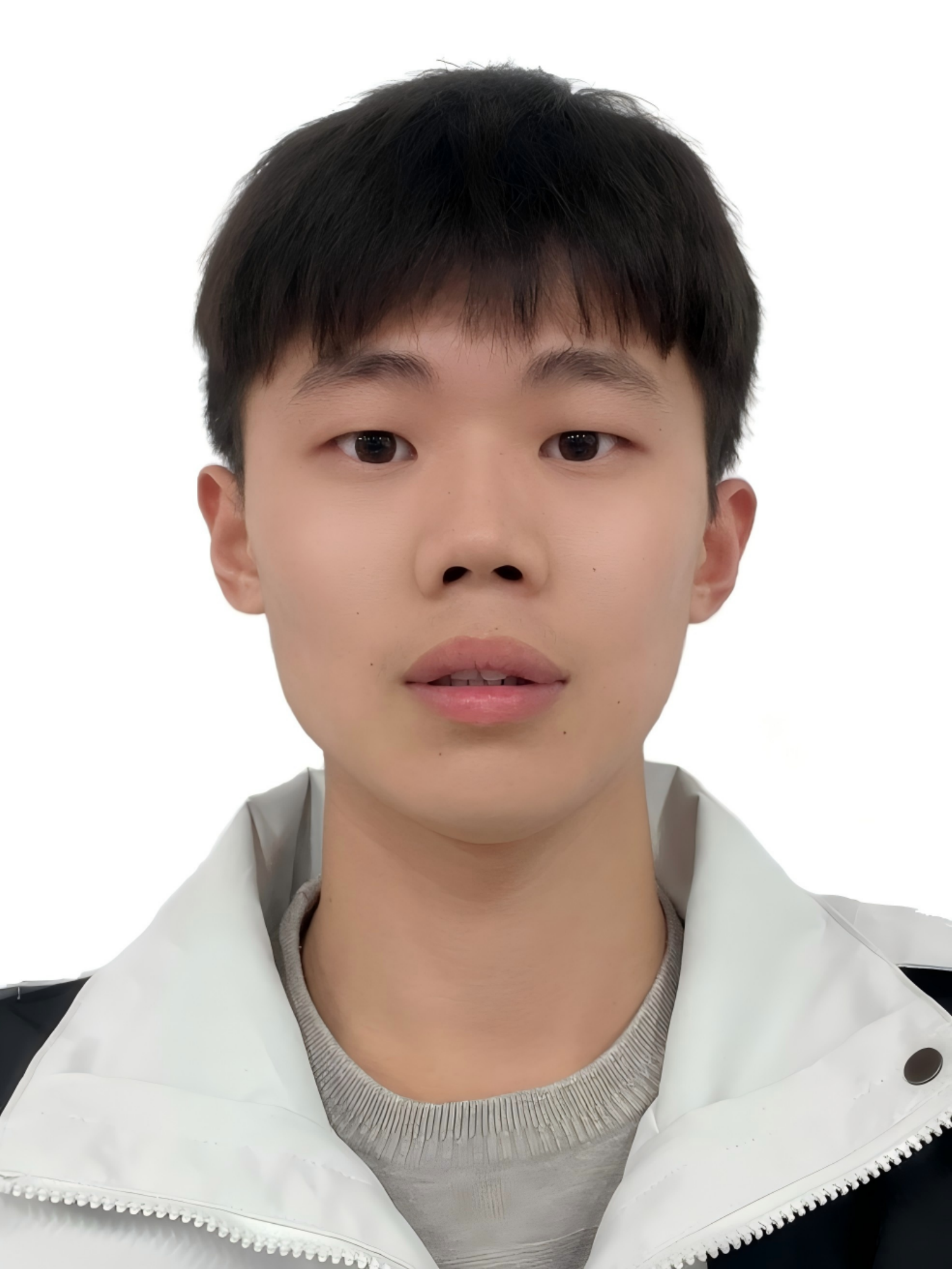}}]{Erhu Liu} received the B.S. and M.S. degrees from Anhui University of Science and Technology, Huainan, China, in 2021 and 2024, respectively. He is currently pursuing a Ph.D. degree in computer science and technology with Chongqing University of Posts and Telecommunications, Chongqing, China. His research interests include the adversarial robustness of machine learning and deep learning.
\end{IEEEbiography}

\vspace{-1em}
\begin{IEEEbiography}[{\includegraphics[width=0.91in,height=1.25in,clip,keepaspectratio]{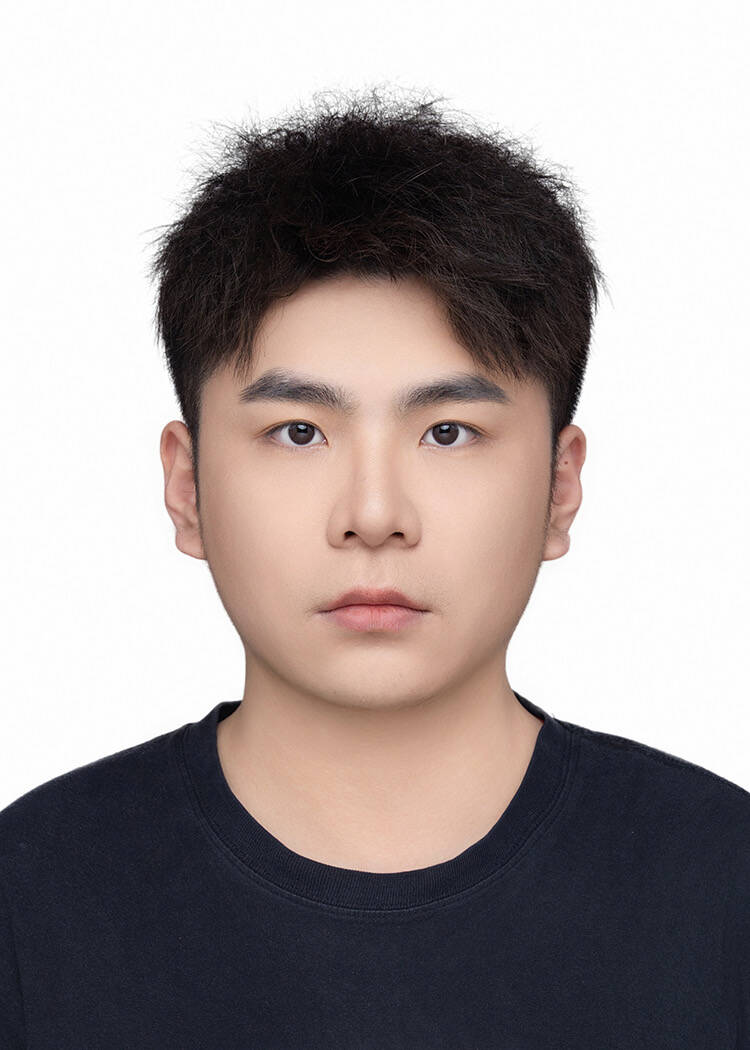}}]{Zonglin Yang} received the B.S. degree in networks engineering from Hunan Institute of Technology in 2021 and the M.S. degree in computer science and technology  from Chongqing University of Posts and Telecommunications in 2024. He is currently pursuing the Ph.D. degree in Biological Systems Engineering with University of Nebraska-Lincoln, NE, USA. His research interests include image processing, multimedia security and forensics.
\end{IEEEbiography}

\vspace{-1em}
\begin{IEEEbiography}[{\includegraphics[width=0.91in,height=1.25in,clip,keepaspectratio]{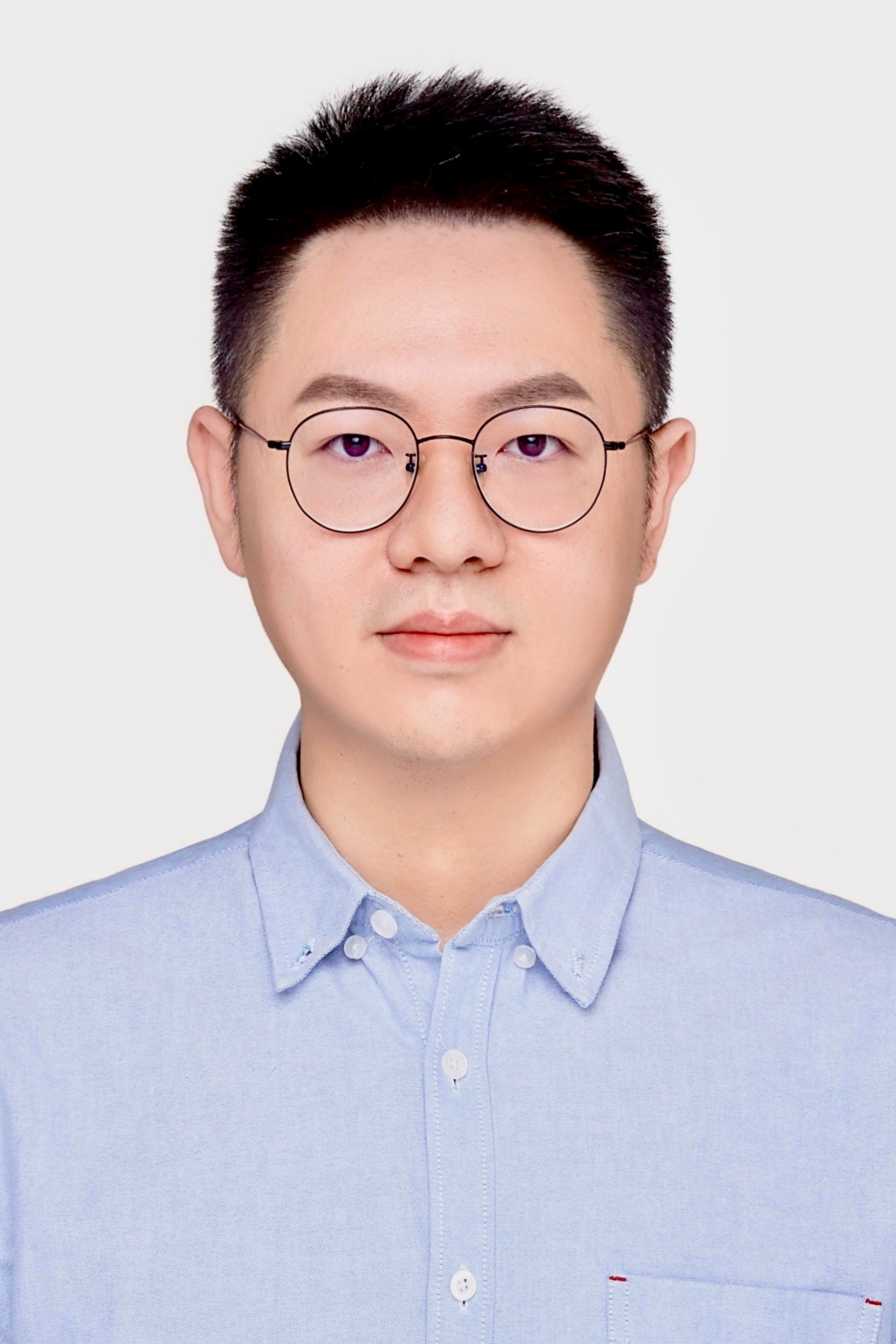}}]{Bo Liu} (Member, IEEE) received the B.B.M. degree from the Department of Information Technology, University of Internation Relations, China, in 2010, the M.S. and Ph.D. degree from the Department of Computer and Information Science, University of Macau, Macao S.A.R., China, in 2014, and 2020, respectively. He is currently a Lecturer with the School of Computer Science and Technology, Chongqing University of Posts and Telecommunications, China. His current research interests include multimedia forensics, image processing, and computer vision. 
\end{IEEEbiography}

\vspace{-1em}
\begin{IEEEbiography}[{\includegraphics[width=0.91in,height=1.25in,clip,keepaspectratio]{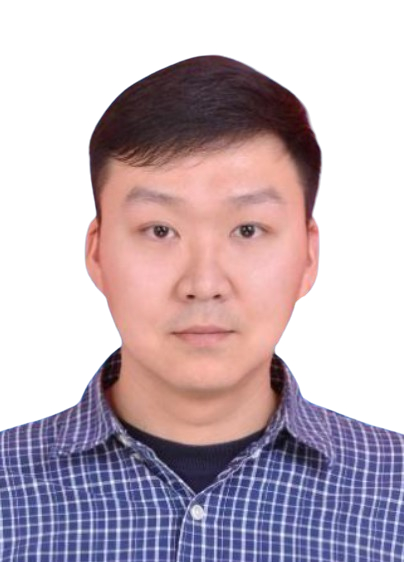}}]{Xianjia Meng} received the Ph.D. degree from Xidian University, Xi’an, China. He is currently a professor with the School of Computer Science, Northwest University, Xi’an. His research interests include parallel and distributed computing, machine learning, and network security. 
\end{IEEEbiography}

\vspace{-1em}
\begin{IEEEbiography}[{\includegraphics[width=0.91in,height=1.25in,clip,keepaspectratio]{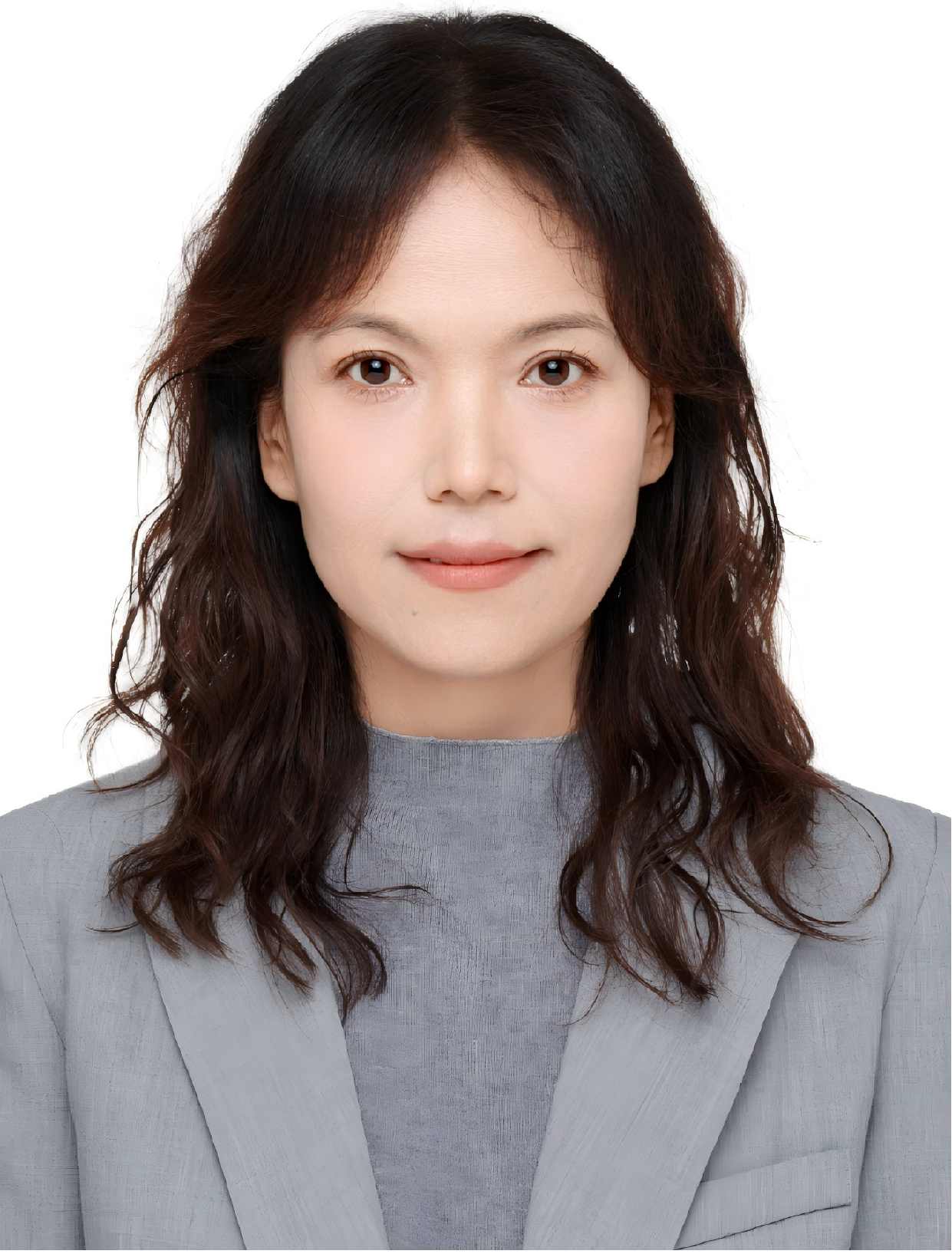}}]{Xiuli Bi} received the B.S. and M.S. degrees from Shanxi Normal University, China, in 2004 and 2007, respectively, and Ph.D. degree in computer science from the University of Macau, in 2017. She is currently a Professor with the College of Computer Science and Technology, Chongqing University of Posts and Telecommunications, China. Her research interests include image processing, multimedia security, and machine learning security.
\end{IEEEbiography}

\newpage
\begin{IEEEbiography}[{\includegraphics[width=0.91in,height=1.25in,clip,keepaspectratio]{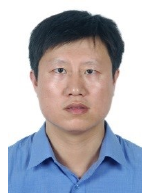}}]{Junwei Han} received the Ph.D. degree from Northwestern Polytechnical University, in 2003. He is a professor with Northwestern Polytechnical University, Xi’an, China. He was a research fellow with Nanyang Technological University, Singapore, The Chinese University of Hong Kong, Hong Kong, and University of Dundee, Dundee, United Kingdom. His research interests include computer vision and brain imaging analysis. He has published more than
100 papers in IEEE Transactions and top tier conferences. He is currently an associate editor of IEEE Transactions on Neural Networks and Learning Systems, IEEE Transactions on Circuits and Systems for Video Technology, and IEEE Transactions on Multimedia.
\end{IEEEbiography}

\newpage
\begin{IEEEbiography}[{\includegraphics[width=0.91in,height=1.25in,clip,keepaspectratio]{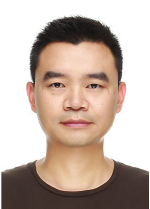}}]{Bin Xiao} received the B.S. and M.S. degrees in electrical engineering from Shanxi Normal University, Xi'an, China, in 2004 and 2007, respectively, and the Ph.D. degree in computer science from Xidian University, Xi'An. He is currently a Professor with Chongqing University of Posts and Telecommunications, Chongqing, China. His research interests include image processing and pattern recognition.
\end{IEEEbiography}

\end{document}